\documentclass[a4paper,fleqn]{cas-sc}

\usepackage[authoryear]{natbib}
\usepackage{booktabs}
\usepackage{amsmath}
\usepackage{amssymb}
\usepackage{amsthm}
\usepackage{graphicx}
\usepackage{float}
\usepackage{tabularx}
\usepackage{longtable} 
\usepackage{booktabs}
\newtheorem{observation}{Observation}
\newtheorem{lemma}{Lemma}
\newtheorem{corollary}{Corollary}
\newtheorem{definition}{Definition}

\newcommand{\nScreened}{23}
\newcommand{\nClusters}{10}

\newcommand{\nFullGrid}{4}
\newcommand{\nObjectives}{7}

\newcommand{\WikiLabels}{260}

\newcommand{\OhsumedLabels}{1{,}986}

\newcommand{\EurlexLabels}{35}

\newcommand{\nEstimableCorpora}{six}
\newcommand{\nEstimableClusters}{four}

\newcommand{\FocalDRel}{+0.037}
\newcommand{\FocalTRel}{4.7}
\newcommand{\FocalTRI}{6.0}

\newcommand{\DistBalDRel}{-0.156}
\newcommand{\DistBalTRel}{4.0}
\newcommand{\DistBalTRI}{0.8}
\newcommand{\DistBalTRS}{5.7}

\newcommand{\OhsumedPooledRatio}{9.70}

\newcommand{\AmazonGateRel}{0.297}

\newcommand{\AmazonSubRel}{0.474}

\newcommand{\AmazonDerivSigma}{2.9}
\newcommand{\AmazonDerivDelta}{+0.177}

\newcommand{\OhsumedWithinRatio}{1.47}

\newcommand{\ChileanLabels}{24}
\newcommand{\ChileanTail}{12}
\newcommand{\ChileanEstimable}{3}
\newcommand{\ChileanNoCompanion}{9}

\newcommand{\EstimatorBiasStrongWorst}{-0.19}

\newcommand{\EstimatorStrengthFloor}{0.8}

\newcommand{\EurlexEffectBce}{+0.280}
\newcommand{\EurlexEffectBceCI}{0.199}

\newcommand{\WikiEffectBce}{+0.133}

\newcommand{\OhsumedEffectBce}{+0.131}
\newcommand{\OhsumedEffectBceCI}{0.014}

\newcommand{\ScumbleR}{+0.694}
\newcommand{\ScumbleLblR}{+0.590}
\newcommand{\ScumbleNCorpora}{23}

\newcommand{\nObjectiveCells}{18}

\newcommand{\nObjectiveTested}{18}
\newcommand{\nObjectiveCalled}{5}
\newcommand{\nObjectiveHolm}{1}
\newcommand{\LlmDocuments}{9{,}986}
\newcommand{\LlmTestDocuments}{293{,}855}
\newcommand{\LlmCoverage}{3.4}
\newcommand{\LlmLabels}{111}
\newcommand{\LlmEffect}{+0.0656}
\newcommand{\LlmRelEffect}{0.297}
\newcommand{\LlmBceEffect}{+0.1429}
\newcommand{\LlmBceRelEffect}{0.440}

\newcommand{\CornetEurlexControlA}{0.654}
\newcommand{\CornetEurlexControlB}{0.561}
\newcommand{\CornetEurlexControlSpread}{0.092}
\newcommand{\CornetEurlexShift}{+0.011}

\newcommand{\LadEurlexBase}{0.591}
\newcommand{\LadEurlexBlind}{1.228}
\newcommand{\LadEurlexBlindSigma}{4.0}

\newcommand{\LadOhsumedBase}{0.367}
\newcommand{\LadOhsumedBlind}{0.717}
\newcommand{\LadOhsumedBlindSigma}{6.8}

\newcommand{\LadPubmedGate}{0.319}
\newcommand{\LadPubmedSubst}{0.736}
\newcommand{\LadPubmedSubstSigma}{2.6}
\newcommand{\LadWikiBase}{0.960}
\newcommand{\LadWikiBlind}{0.312}

\newcommand{\LadWikiGate}{0.941}
\newcommand{\LadWikiSubst}{0.547}

\newcommand{\CorrectiveMinSeeds}{5}
\newcommand{\CorrectiveMaxDeltaRI}{0.028}
\newcommand{\CorrectiveMaxDeltaRISigma}{0.9}

\newcommand{\InflEEurlex}{0.141}

\newcommand{\InflExcessPubmed}{7.6}

\newcommand{\InflOutFSharePubmed}{33}

\newcommand{\InflOhsumed}{0.025}
\newcommand{\InflPiTestOhsumed}{0.189}
\newcommand{\InflPiBarOhsumed}{0.570}
\newcommand{\InflVacuousOhsumed}{47}
\newcommand{\InflRelOhsumed}{0.156}
\newcommand{\InflExcessOhsumed}{18.4}

\newcommand{\InflVacuousWiki}{53}

\newcommand{\InflExcessWiki}{3.0}

\newcommand{\InflEAmazon}{0.047}

\newcommand{\InflNCorpora}{8}

\newcommand{\BootCells}{18}
\newcommand{\BootExclude}{2}
\newcommand{\BootFlagged}{5}
\newcommand{\BootFlaggedExclude}{2}
\newcommand{\SurvRecallDelta}{-0.010}
\newcommand{\SurvRecallSnr}{-1.1}

\newcommand{\CaseLabel}{\emph{periapical diseases}}
\newcommand{\CaseCompanions}{\emph{adolescence}, \emph{adult}, \emph{child}, \emph{female}, \emph{follow-up studies}}
\newcommand{\CaseHitsS}{11}
\newcommand{\CaseNS}{17}
\newcommand{\CaseHitsI}{0}
\newcommand{\CaseNI}{10}

\begin{document}

\shorttitle{When the tail is earned by its companions}
\shortauthors{C. Löffler}

\title[mode=title]{When the tail is earned by its companions: measuring label dependence in long-tailed multi-label classification}

\author[1]{Christoffer Löffler}[orcid=0000-0003-1834-8323]
\cormark[1]
\ead{christoffer.loffler@pucv.cl}
\affiliation[1]{organization={School of Computer Engineering, Pontificia Universidad Católica de Valparaíso},
    addressline={Av. Brasil 2950},
    city={Valparaíso},
    postcode={2340025},
    state={Valparaíso},
    country={Chile}}
\cortext[1]{Corresponding author}

\begin{abstract}
A rare label seldom occurs alone, and a classifier can recover it from the frequent labels it co-occurs with, its companions, without having learned the label itself. Existing diagnostics of this need a trained model: none says before training whether a corpus admits the failure, or what the standard aggregate hides. Macro-F1, like any measure built from per-label confusion counts, is unchanged by how a label's recall divides between occurrences with and without a companion. We introduce a within-label estimand of that division, a reading rule under which a fall in dependence counts only if companion-free recall rises, and a statistic of training labels that bounds, before training, how far reported recall can exceed companion-free recall. Across \nScreened{} corpora, \nObjectives{} training objectives and five families of label-correlation mechanisms, companion-free recall falls short on all \nEstimableCorpora{} corpora where the estimand is defined. Up to \InflExcessOhsumed\% of the tail true positives macro-F1 credits would vanish at the companion-free rate, within the training-label bound on every corpus. No long-tail objective reduces the dependence, but architecture moves it. Models whose tail score is computed from other labels' predictions are strongly dependent.
\end{abstract}


\maketitle

\section{Introduction}
\label{sec:introduction}

Long-tailed multi-label classification is usually framed as a difficulty of learning: most labels are rare, rare labels are hard~\citep{zhangReviewMultiLabelLearning2014,tarekegnReviewMethodsImbalanced2021}, and the field's methods address that difficulty with reweighted losses~\citep{linFocalLossDense2018,cuiClassBalancedLossBased2019,wuDistributionBalancedLossMultiLabel2021,ben-baruchAsymmetricLossMultiLabel2021}, margin adjustments~\citep{caoLearningImbalancedDatasets2019,menonLongtailLearningLogit2021a} and architectures that exploit label correlation~\citep{chenMultiLabelImageRecognition2019,xunCorrelationNetworksExtreme2020,yangSGMSequenceGeneration2018,yanLabelCoRankRevolutionizingLong2025a}. Progress is measured by macro-F1 over the tail~\citep{xiaoDoesHeadLabel2021,xiaoPairwiseInstanceRelation2022,xuLabelSpecificFeatureAugmentation2023}, which weights every label equally and so refuses to let frequent labels dominate the score.

Rare labels are not merely rare, however. They are rare and they seldom occur alone. In large annotation vocabularies a tail label often appears in documents that also carry a few frequent labels~\citep{charteConcurrenceImbalancedLabels2014,xuLabelSpecificFeatureAugmentation2023}, and those frequent labels are easy. A classifier can therefore recover the rare label reliably without having learned anything about it, by recognizing its frequent companions and predicting the rare label alongside them~\citep{geirhosShortcutLearningDeep2020,fanAccurateUseLabel2023}. Where the rare label appears without its companions the same classifier may fail entirely~\citep{singhDontJudgeObject2020,xieCounterfactualReasoningMultiLabel2024}.

What is missing is a measurement. The phenomenon itself is known: models amplify the co-occurrence statistics of their training corpora~\citep{zhaoMenAlsoShopping2017}, co-occurring classes are exploited as a shortcut in vision~\citep{singhDontJudgeObject2020,xieCounterfactualReasoningMultiLabel2024} and text~\citep{fanAccurateUseLabel2023}, and many methods use label correlation deliberately as a feature~\citep{chenMultiLabelImageRecognition2019,xunCorrelationNetworksExtreme2020}. Each of these works either removes the dependence or exploits it, and the diagnostics they use are computed from a trained model's errors. None can say before training whether a corpus admits the failure, which is when a benchmark is chosen and compute is committed, and none relates the dependence to what an evaluation measure reports. Statistics of label co-occurrence in datasets exist~\citep{herreraMultilabelClassification2016,charteConcurrenceImbalancedLabels2014}, but none is tied to what an evaluation measure can hide, and the literature on the blind spots of evaluation measures~\citep{sokolovaSystematicAnalysisPerformance2009,wuUnifiedViewMultiLabel2017} has not addressed this one. This paper supplies the instrument: a measure of the dependence, a characterization of what the aggregate cannot see, and a bound available before training.

Two models with identical macro-F1 can differ completely in whether they recover a rare label alongside its companions or without them. Per-label F1, and therefore any mean of per-label F1 scores, depends only on how many of a label's occurrences are recovered, not on which, so it is unchanged by how recall divides between the occurrences that carry a companion (supported) and those that carry none (independent). The same holds for every measure built from per-label confusion counts.

We therefore measure the division directly. The within-label effect $d$ is the mean, over tail labels, of recall on supported occurrences minus recall on independent ones. It is read under a co-report rule: a fall in $d$ counts as a correction only if independent recall rises, since a model can lower $d$ by predicting less on the supported occurrences. A screening statistic, the mean support ratio $\bar{\pi}$, is computed from training labels alone and bounds the inflation of reported recall, $R - R_I$, the amount by which a label's recall exceeds its recall on independent occurrences. We then audit \nScreened{} corpora in \nClusters{} clusters with \nObjectives{} training objectives and five families of correlation-exploiting mechanisms. The mechanisms divide into those that only adjust a tail label's document-based score with label correlation and those whose tail score is derived from other labels' predictions. The stakes are concrete: on Ohsumed, \InflExcessOhsumed\% of the tail true positives that macro-F1 credits would vanish if companion-supported occurrences were recalled at the companion-free rate, and between \InflExcessWiki\% and \InflExcessPubmed\% on the other large-vocabulary corpora.

We use the following running example: on Ohsumed, the MeSH descriptor \CaseLabel{} has the companions \CaseCompanions{}. A binary cross-entropy model recovers it \CaseHitsS{} times out of \CaseNS{} on test documents that carry one of those companions and \CaseHitsI{} times out of \CaseNI{} on documents that carry none. Macro-F1 records the \CaseHitsS{} hits and nothing about where they fell. Table~\ref{tab:glossary} collects the vocabulary used throughout.

\begin{table}[t]
\caption{Vocabulary used throughout the paper.}
\label{tab:glossary}
\small
\begin{tabular*}{\tblwidth}{@{}lp{0.72\linewidth}@{}}
\toprule
Term & Meaning \\
\midrule
companion $C(l)$ & a frequent (head) label that co-occurs with tail label $l$ above a threshold, estimated on training labels (Definition~\ref{def:companion}) \\
supported, independent & an occurrence of $l$ that does, or does not, carry a member of $C(l)$ (Definition~\ref{def:partition}) \\
support ratio $\pi(l)$ & the share of $l$'s occurrences that are supported; $\bar{\pi}$ is its mean over the tail \\
$R_S$, $R_I$ & recall on supported and on independent occurrences \\
within-label effect $d$ & the mean over estimable tail labels of $R_S(l) - R_I(l)$ (Equation~\ref{eq:within}) \\
inflation & $R(l) - R_I(l)$, how far reported recall exceeds independent recall; bounded by $\pi(l)$ \\
co-report & reading rule under which a fall in $d$ counts as a correction only if $R_I$ rises (Definition~\ref{def:damage}) \\
arm, control & a trained model variant, and the variant identical to it but for one mechanism \\
adjusting, deriving & an arm that modifies a document-based tail score with label correlation, or one that computes the tail score from other labels' predictions (Section~\ref{sec:arms}) \\
\bottomrule
\end{tabular*}
\end{table}

\subsection{Research objectives}
\label{sec:rq}

The study answers four questions.

\begin{enumerate}
\item[\textbf{RQ1}] Do trained classifiers recover tail labels more often alongside their companions than without them, and on which corpora can that be measured? (Sections~\ref{sec:validity} and~\ref{sec:establish})
\item[\textbf{RQ2}] Do standard long-tail training objectives reduce companion dependence? (Section~\ref{sec:objectives})
\item[\textbf{RQ3}] How much of it can be known from training labels alone, before any model is trained? (Section~\ref{sec:screening-results})
\item[\textbf{RQ4}] Which design property of a model moves it? (Section~\ref{sec:derive})
\end{enumerate}

Our contributions are as follows.

\begin{enumerate}
\item A training-free screening statistic, the mean support ratio $\bar{\pi}$, computed from training labels alone, that bounds before any model is trained how far reported tail recall can exceed companion-free recall. Existing diagnostics of companion dependence need a trained model. Our diagnostic tells, when a benchmark is chosen, whether a corpus admits the failure at all.
\item A within-label estimand with a co-report rule, which reads a fall in dependence as a correction only if companion-free recall rises. Unlike the intuitive difference of F1 scores, it is not driven by base rates and cannot be lowered by damaging the model.
\item An exact decomposition of per-label recall into companion-supported and independent components, under which every measure built from per-label confusion counts is invariant to how recall divides between them. This invariance motivates the estimand. We characterize how wide the blind spot is in practice.
\item An audit of \nScreened{} corpora, of which the estimand is defined on \nEstimableCorpora{} (\nEstimableClusters{} independent clusters), across training objectives and mechanism families. Its findings locate the dependence: the training objective does not move it, but whether a tail score is derived from other labels' predictions does.
\item A model-agnostic evaluator that reads prediction files, used unchanged on a retrieval-augmented language model.
\end{enumerate}

Section~\ref{sec:related} reviews related work. Section~\ref{sec:formalism} gives the formalism: the invariance, the within-label estimand, the co-report rule and the screening statistic. Section~\ref{sec:setup} describes datasets, models, mechanism arms and protocol. Section~\ref{sec:results} answers RQ1--RQ4 and applies the evaluator to a system that is not a classifier. Section~\ref{sec:discussion} discusses implications and limitations, and Section~\ref{sec:conclusion} concludes.

\section{Related work}
\label{sec:related}

Label dependence in multi-label learning has been analyzed theoretically by \citet{dembczynskiLabelDependenceLoss2012}, who distinguish conditional from marginal dependence and use the distinction to choose learning objectives. \citet{singhDontJudgeObject2020} separate ``exclusive'' from ``co-occurring'' test distributions in object recognition, report accuracy on each, and propose de-correlating training objectives. Both works concern models and objectives rather than measurement.

The closest line of work to our screening statistic is SCUMBLE and REMEDIAL~\citep{charteConcurrenceImbalancedLabels2014,charteResamplingMultilabelDatasets2015,charteDealingDifficultMinority2019}. SCUMBLE measures the concurrence of minority with majority labels without training, and its per-label variant SCUMBLE$_{Lbl}$~\citep{charteDealingDifficultMinority2019} is computed from the label matrix alone. Its purpose is to estimate how hard a dataset is to resample, and REMEDIAL decouples training instances on the strength of the statistic~\citep{charteResamplingMultilabelDatasets2015}. The line of work neither audits datasets with an evaluation measure nor has a notion of entailment: a companion that accompanies a label by definition counts the same as one whose association is behavioral. Over the \ScumbleNCorpora{} corpora of our study, SCUMBLE$_{Lbl}$ correlates with our screening ratio $\bar{\pi}$ at $r = \ScumbleLblR$ (dataset-level SCUMBLE: $\ScumbleR$). The two measure neighboring quantities by different constructions: SCUMBLE is an inequality index over per-label imbalance ratios, $\bar{\pi}$ a proportion of a label's positives.

\subsection{The phenomenon as a shortcut}

A model that succeeds by a feature that does not generalize is learning a shortcut~\citep{geirhosShortcutLearningDeep2020}. Companion dependence is an instance. The companion helps predict the tail label during training and is absent when the tail label later appears alone. \citet{zhaoMenAlsoShopping2017} compute a corpus-level bias score from training co-occurrence counts and show that models amplify it at test time, a fairness harm motivating corpus-level correction. Our statistic instead bounds what an evaluation measure can conceal. \citet{gardnerCompetencyProblemsFinding2021} argue that for a class of language tasks every feature correlation is spurious and derive a statistical test for artifacts from data alone. Our screening statistic is similar in spirit, restricted to co-occurrence between labels and tied to a specific claim about what an aggregate score can hide.

\citet{xieCounterfactualReasoningMultiLabel2024} treat co-occurring features as carrying both a helpful and a harmful causal effect and diagnose the harm with a conditional false-positive rate computed from a trained model's errors. \citet{fanAccurateUseLabel2023} name the mechanism in text the label correlation shortcut and propose a counterfactual classifier to remove it, and \citet{rawlekarImprovingMultilabelRecognition2024} enforce class co-occurrence probabilities through a graph network. None of these supplies a statistic available before training. \citet{xuLabelSpecificFeatureAugmentation2023} identify companion dependence in long-tailed multi-label text classification and observe that augmenting data for one label also affects its co-occurring labels, \citet{xiaoPairwiseInstanceRelation2022} augment tail instances pairwise, \citet{lanchantinGeneralMultilabelImage2020} mask labels so a model learns from partial label sets, and \citet{duartePLMPartialLabel2021} mask labels to correct class imbalance rather than correlation. \citet{xiaoDoesHeadLabel2021} meta-learn a mapping from head-label to tail-label classifiers, and \citet{wangValueHeadLabels2024} extend it with model- and feature-level knowledge from head labels. These transfer parameters rather than predictions and so modify, rather than replace, a document-based tail score.

Every treatment above either reduces the dependence or exploits it. None is an instrument, i.e., supplies a measure that quantifies the dependence, a characterization of what the standard aggregate cannot see, or a bound available before training on how far the aggregate can be inflated.

\subsection{Group robustness and dataset characterization}

The paper is closely related to worst-group evaluation under spurious correlation, where an average conceals a group on which the model fails. \citet{oakden-raynerHiddenStratificationCauses2019} name this hidden stratification, and \citet{sagawaDistributionallyRobustNeural2020} compute worst-group accuracy with groups given by annotated attributes. Methods for unannotated subgroups infer them by clustering a trained model's features~\citep{sohoniNoSubclassLeft,eyubogluDOMINODISCOVERINGSYSTEMATIC2022}. In multi-label data the groups are induced by label co-occurrence and computable from the label matrix alone, which is what makes a screening statistic possible.

A recent survey focused on single-label data~\citep{yeCleverHansMirage2025} notes that available metrics require a trained model. Surveys of multi-label data characterize datasets thoroughly but stop short of a metric. \citet{herreraMultilabelClassification2016} catalog cardinality, density, mean imbalance ratio, SCUMBLE and the theoretical complexity score (TCS), and \citet{tarekegnReviewMethodsImbalanced2021} survey the imbalanced multi-label methods built on them. \citet{zhangReviewMultiLabelLearning2014} organize methods by how much label correlation they model, an axis \citet{huangApplicationLabelCorrelation2024} restate for current work, and \citet{szymanskiHowDataDrivenApproach2016} build label co-occurrence graphs from training data to partition the label space for method design. These characterizations carry no evaluation measure.

\subsection{Properties of evaluation measures}
\label{sec:metric-theory}

Asking what an evaluation measure cannot see is a well-studied question. \citet{sokolovaSystematicAnalysisPerformance2009} give the canonical taxonomy of which changes to a confusion matrix leave which classification measures unchanged. For multi-label measures, \citet{wuUnifiedViewMultiLabel2017} unify eleven metrics through label-wise and instance-wise margins, \citet{sebastianiAxiomaticallyDerivedMeasure2015} derives a measure from axioms and shows that F1 violates several of them, \citet{waegemanBayesoptimalityFmeasureMaximizers2015} show that optimizing Hamming or subset 0/1 loss as a surrogate for the F-measure incurs high worst-case regret, \citet{opitzCloserLookClassification2024} observes that macro-averaged measures are routinely chosen without their properties being stated, and \citet{pereiraCorrelationAnalysisPerformance2018} find most multi-label measures highly correlated with one another. The invariance studied here is of a different kind from those in the taxonomy of \citet{sokolovaSystematicAnalysisPerformance2009}, which concern changes a confusion matrix can express. Section~\ref{sec:invariance} returns to the comparison.

Information retrieval faces a structurally similar problem when the evaluation signal is partial. \citet{buckleyRetrievalEvaluationIncomplete2004} design bpref for incomplete relevance judgments, computing it only over judged documents, an idea our estimand takes up. \citet{sakaiAlternativesBpref2007} shows that standard measures applied to condensed lists are more robust to incompleteness than bpref itself. \citet{ferranteHowIntervalScales2020} determine which arithmetic a measure permits, \citet{amigoComparisonExtrinsicClustering2009} derive formal constraints an evaluation metric should satisfy, and \citet{fuhrCommonMistakesIR2018} catalogs recurring errors in retrieval evaluation, several of which stem from inadequate measures.

\subsection{Propensity-scored evaluation}

Propensity-scored precision~\citep{jainExtremeMultilabelLoss2016} corrects for labels missing from the ground truth. Our measure is orthogonal, in that it addresses predictions that are present in the ground truth but correct for the wrong reason, in the sense of \citet{mccoyRightWrongReasons2019} for natural language inference, where a model scores well through a heuristic that holds for frequent cases and breaks on the rest. Propensity-scored precision conflates missing labels with the long tail~\citep{schultheisMissingLabelsLongtails2022}. Its scores can exceed their own maximum and its fitted parameters do not transfer between collections. This led to budgeted utilities that target long-tail performance directly~\citep{schultheisGeneralizedTestUtilities2024} and to consistent algorithms for the macro-at-$k$ family~\citep{schultheisConsistentAlgorithmsMultilabel2024}. Whether a correct prediction was made for the right reason is a third axis beside missingness and tail weighting, and the replacement metrics are themselves built from per-label confusion counts, so correcting the tail weighting does not recover the split within a label. \citet{weiDoesTailLabel2020} make the complementary observation that tail labels affect Top-$k$ precision and nDCG@$k$ so little that they can be trimmed from the label space with little loss on either, and \citet{ullahLabelsExtremesHow2024} find that standard extreme-classification evaluation, which captures only the order of scores, says nothing about calibration.

\section{Formalism}
\label{sec:formalism}

This section fixes notation, characterizes what count-based measures cannot see, and derives our estimand. Two constructions for it suggest themselves, and both are confounded, one by a base-rate artifact severe enough to invert its sign and one by Simpson's paradox. The estimand instead measures the effect within each label and averages, and it is read under a rule that a reduction in dependence counts only when independent recall rises.

\subsection{Setting}
\label{sec:setting}

Let $L$ be a label set, $\mathcal{X}$ a set of documents, and $Y(x) \subseteq L$ the gold labels of $x \in \mathcal{X}$. A classifier predicts $\hat{Y}(x) \subseteq L$. Write $n_l = |\{x : l \in Y(x)\}|$ for the frequency of label $l$. A prevalence threshold $\theta$ partitions $L$ into a head set $H$ and a tail set $T$. Both $\theta$ and the companion threshold below are parameters and both are swept. Section~\ref{sec:protocol} gives the reported values and Appendix~\ref{app:sensitivity} the sweep.

\begin{definition}[Companion set]
\label{def:companion}
For a tail label $l \in T$ and a head label $h \in H$, write
\begin{equation*}
p(h \mid l) = P\bigl(h \in Y(x) \mid l \in Y(x)\bigr),
\qquad
\mathrm{co}(h,l) = \bigl|\{x : \{h,l\} \subseteq Y(x)\}\bigr|,
\end{equation*}
where probabilities are empirical frequencies over the training split. Given a companion threshold $\tau \in [0,1]$, a minimum lift $\lambda$ and an entailment cutoff $c$,
\begin{equation}
C(l) = \Bigl\{h \in H : p(h \mid l) \geq \tau,\quad
\frac{p(h \mid l)}{P(h \in Y(x))} \geq \lambda,
\quad p(h \mid l) < c, \quad \mathrm{co}(h,l) > 0 \Bigr\}.
\label{eq:companion}
\end{equation}
We use $\lambda = 1$ and $c = 1$ throughout and suppress the dependence of $C(l)$ on $(\tau, \lambda, c)$.
\end{definition}

The conditional runs head-given-tail, because $P(l \mid h)$ is vanishingly small for every rare $l$ and would select nothing. The requirement $\mathrm{co}(h,l) > 0$ matters only at $\tau = 0$, where it restricts $C(l)$ to head labels that actually co-occur with $l$. The other two conditions change what is measured, and both are set to their weakest admissible values. The lift condition controls for the head label's base rate. Without it, a head label occurring in a fraction $\geq \tau$ of all documents is a companion of every tail label, associated or not. At $\lambda = 1$ it requires only that the head be no rarer given $l$ than overall. The entailment cutoff excludes head labels that the annotation scheme guarantees, a concern shared by \citet{amigoEvaluatingExtremeHierarchical2022} in discounting hierarchically implied labels. At $c = 1$ it removes only exceptionless companions, those with $p(h \mid l) = 1$. On a hierarchically closed vocabulary, where every label entails its ancestors, omitting it makes every positive of every tail label supported by construction and leaves the partition below degenerate.

\begin{definition}[Supported and independent occurrences]
\label{def:partition}
For a tail label $l \in T$, the positives of $l$ partition into
\begin{align}
S(l) &= \{x : l \in Y(x),\ Y(x) \cap C(l) \neq \emptyset\}, \\
I(l) &= \{x : l \in Y(x),\ Y(x) \cap C(l) = \emptyset\},
\end{align}
the \emph{supported} and \emph{independent} occurrences of $l$, with $|S(l)| + |I(l)| = n_l$. Write $\pi(l) = |S(l)| / n_l$ for the support ratio and
\begin{equation*}
R_S(l) = \frac{|\{x \in S(l) : l \in \hat{Y}(x)\}|}{|S(l)|},
\qquad
R_I(l) = \frac{|\{x \in I(l) : l \in \hat{Y}(x)\}|}{|I(l)|}
\end{equation*}
for recall within each context, each undefined where its denominator is zero.
\end{definition}

The entailment cutoff is crucial in practice, for example, on arXiv-157 with every label's ancestors added (the register's ``arXiv-157 hier.''), it moves $\bar{\pi}$ from $0.845$ to $0.024$, so almost all of the apparent companion structure there is entailment rather than behavior. This is also what separates the construction from the concurrence statistics of Section~\ref{sec:related}, which count a definitional companion the same as a behavioral one.

The companion set $C(l)$ is estimated once from the training labels and held fixed. The partition is then applied to whichever split is under discussion, and $n_l$, $S(l)$, $I(l)$ and $\pi(l)$ denote counts on that split. This is what makes $\pi$ available before a model exists; re-estimating $C(l)$ per split would make the partition a function of the evaluation data. Because $I(l)$ is the complement of $S(l)$ among the positives of $l$, recall decomposes as an identity,
\begin{equation}
R(l) = \pi(l) \, R_S(l) + \bigl(1 - \pi(l)\bigr) R_I(l),
\label{eq:decomposition}
\end{equation}
wherever both terms are defined, that is, wherever $l$ has a positive in each context.

Two aggregates of $\pi(l)$ appear below. The screening statistic
\begin{equation}
\bar{\pi} = \frac{1}{|T|} \sum_{l \in T} \pi(l)
\label{eq:pibar}
\end{equation}
is the unweighted mean over the tail, computed on training labels. The pooled ratio
\begin{equation}
\pi_{\text{pool}} = \frac{\sum_{l \in T} |S(l)|}{\sum_{l \in T} n_l}
\label{eq:pipool}
\end{equation}
weights each label by its frequency and is computed on the test split, beside an effect measured on that split. The two can differ substantially, reading $0.570$ and $0.147$ on Ohsumed, and every value we report states which it is.

\subsection{What count-based measures cannot see}
\label{sec:invariance}

\begin{definition}[Context redistribution]
\label{def:redistribution}
A \emph{context redistribution} of a tail label $l$ is any change from $\hat{Y}$ to $\hat{Y}'$ that alters which positives in $S(l) \cup I(l)$ satisfy $l \in \hat{Y}(x)$ while leaving their number unchanged, and leaves $\hat{Y}$ unchanged on every document with $l \notin Y(x)$ and on every label other than $l$.
\end{definition}

\begin{observation}[Invariance]
\label{prop:invariance}
Every evaluation measure that is a function of the per-label confusion counts $(\mathrm{TP}_{l}, \mathrm{FP}_{l}, \mathrm{FN}_{l})_{l \in L}$ is unchanged by every context redistribution, while $R_S(l)$ and $R_I(l)$ in general are not.
\end{observation}

The observation is immediate. A redistribution leaves $\mathrm{TP}_l$, and hence $\mathrm{FN}_l = n_l - \mathrm{TP}_l$, unchanged. It leaves predictions on documents without $l$, and hence $\mathrm{FP}_l$, unchanged. And it leaves every other label untouched. Hamming loss, micro-F1 and macro-F1 over any label subset are such measures. They are the three of the eleven measures unified by \citet{wuUnifiedViewMultiLabel2017} that are computed from thresholded prediction sets. Six of the remaining eight are computed from score rankings, to which the definition does not apply as stated. The other two, instance-F1 and instance-AUC, are computed per document and do respond to a redistribution, but through document cardinality. The documents gaining and losing a prediction carry different numbers of labels, so they also respond to redistributions unrelated to the partition. Unlike the invariances in the taxonomy of \citet{sokolovaSystematicAnalysisPerformance2009}, this one concerns a change the confusion matrix cannot express at all, since the per-label counts are identical before and after.

The observation is close to definitional. Its content lies in how much the counts leave undetermined, which the following lemma quantifies.

\begin{lemma}[Reachable split]
\label{lem:reachable}
Holding $R(l)$ and $\pi(l)$ fixed, every pair $(R_S(l), R_I(l))$ satisfying Equation~\ref{eq:decomposition} is reachable by a context redistribution, subject to integrality, and $R_S(l)$ ranges over
\begin{equation}
\Bigl[\max\bigl(0,\ \tfrac{R(l) - (1 - \pi(l))}{\pi(l)}\bigr),\quad
\min\bigl(1,\ \tfrac{R(l)}{\pi(l)}\bigr)\Bigr].
\label{eq:reachable}
\end{equation}
The interval is the whole of $[0,1]$ exactly when $\pi(l) \leq R(l) \leq 1 - \pi(l)$, which requires $\pi(l) \leq 1/2$.
\end{lemma}

\begin{proof}
$\mathrm{TP}_l = n_l R(l)$ constrains only the convex combination $\pi R_S + (1-\pi) R_I$, and a redistribution can move any number of recovered positives between $S(l)$ and $I(l)$ within the bounds $0 \leq R_S, R_I \leq 1$. Solving $\pi R_S + (1-\pi) R_I = R$ for $R_S$ at $R_I = 1$ and $R_I = 0$, and clipping to $[0,1]$, gives the interval.
\end{proof}

\begin{figure}
\centering
\includegraphics[width=0.72\linewidth]{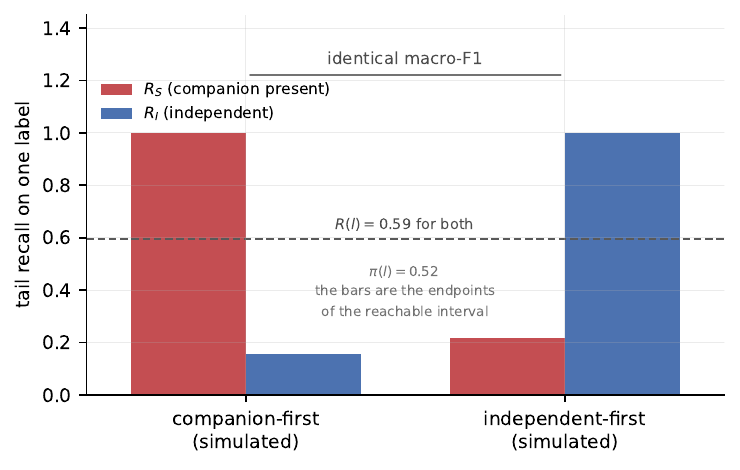}
\caption{The invariance, on a simulated tail label. Two simulated models with the same overall recall $R(l)$, and therefore the same per-label confusion counts and the same macro-F1, split that recall oppositely: one recovers the label mostly alongside a companion, the other mostly without. The two sit at the endpoints of the interval of Lemma~\ref{lem:reachable} at this label's support ratio. No aggregate built from per-label confusion counts distinguishes them.}
\label{fig:proposition}
\end{figure}

Within that range one model recovers the label only alongside a companion and another only without it, and the two report identical confusion counts. Above $\pi(l) = 1/2$ the range narrows, since a label whose positives are mostly supported cannot place all of its recall on the independent ones. Figure~\ref{fig:proposition} draws the two endpoints for a simulated label.

Equation~\ref{eq:decomposition} also gives a bound on the inflation of reported recall over independent recall.

\begin{corollary}[Screening bound]
\label{cor:screening}
For every tail label with $|I(l)| \geq 1$,
\begin{equation}
R(l) - R_I(l) = \pi(l) \bigl(R_S(l) - R_I(l)\bigr) \leq \pi(l),
\label{eq:bound}
\end{equation}
with $\pi(l)$ on the split on which recall is measured, the inequality following from $R_S(l) - R_I(l) \leq 1$ (for $|S(l)| = 0$ both sides are zero). Averaging over the set $F = \{l \in T : |I(l)| \geq 1\}$ of tail labels whose independent recall is defined,
\begin{equation}
\frac{1}{|F|} \sum_{l \in F} \bigl(R(l) - R_I(l)\bigr) \leq \frac{1}{|F|} \sum_{l \in F} \pi(l).
\label{eq:bound-mean}
\end{equation}
\end{corollary}

The bound holds exactly with $\pi$ computed on the evaluation split, with $C(l)$ fixed from training. The screening statistic $\bar{\pi}$ is computed on training labels over the whole tail, so reading it as a bound on test-split inflation requires the support ratio to transfer from training to test. In practice the transfer errs on the conservative side: companion sets estimated on training labels transfer only partly, and the support ratio on the test split is typically lower. Table~\ref{tab:inflation} reports both quantities per corpus, so the transfer can be checked directly. The bound constrains the inflation of the reported number, not the within-label effect $R_S - R_I$, which is bounded only by $1$ whatever $\pi$ is. A label with $\pi(l)$ close to $1$ has a nearly vacuous bound. Table~\ref{tab:inflation} reports the share of such labels.

\subsection{Between-label confounds}
\label{sec:instrument}

A quantity can respond to a context redistribution and still not measure companion dependence, because it responds to other things as well. Two constructions suggest themselves, and both are confounded.

\paragraph{Difference of F1} The first compares tail F1 with tail F1 restricted to independent occurrences. Let $F_1^T$ be micro-F1 over the tail, and $F_1^{T,I}$ the same quantity with true positives and false negatives counted on independent occurrences only,
\begin{equation}
F_1^{T,I} = \frac{2\,\mathrm{TP}_I}{2\,\mathrm{TP}_I + \mathrm{FP}_I + \mathrm{FN}_I},
\qquad
\mathrm{TP}_I = \sum_{l \in T} |\{x \in I(l) : l \in \hat{Y}(x)\}|,
\label{eq:diff-f1}
\end{equation}
with $\mathrm{FN}_I$ analogous and the gap defined as $(F_1^{T,I} - F_1^T)/F_1^T$. The construction fails in two ways. (a) Macro-averaging it is not defined on real test splits, as most tail labels have positives in only one context, so a macro average would be taken over whichever labels happen to survive, which is why the micro form is used. (b) In the micro form, the precision term carries a base-rate artifact. A false positive of $l$ is a document with $l \in \hat{Y}(x)$ and $l \notin Y(x)$. It lies in neither $S(l)$ nor $I(l)$, so it must be assigned to a context, and $\mathrm{FP}_I$ counts those on documents carrying no member of $C(l)$. Among documents without $l$, those carrying no companion outnumber those carrying one by orders of magnitude, because $C(l)$ is small. The independent context therefore absorbs nearly every false positive, its precision falls, and the gap is negative even for a model whose recall is identical in both contexts. Its size is set by the imbalance of the negatives, a property of the corpus, so it neither cancels in comparisons between models nor transfers between corpora.

\paragraph{Ratio of pooled recalls} The second pools $R_S$ and $R_I$ over the tail, as in Equation~\ref{eq:pipool}, and divides them. Pooling weights each label by its occurrence count in each context, so a label contributes different weight to the numerator and the denominator whenever $\pi(l)$ departs from the tail average, and the ratio compares two differently composed populations of labels. Labels that are usually supported, with high $\pi(l)$, are also the more frequent and therefore easier, so the pooled $R_S$ is dominated by easy labels and the pooled $R_I$ by hard ones. This is Simpson's paradox. On Ohsumed, the pooled ratio reads $\OhsumedPooledRatio$ against $\OhsumedWithinRatio$ measured within labels.

Section~\ref{sec:validity} compares our estimand with both constructions on every audited corpus.

\subsection{Within-label effect}
\label{sec:estimand}

The confounds arise from comparing across labels. We therefore measure within each label and average,
\begin{equation}
d = \frac{1}{|E|} \sum_{l \in E} d_l,
\qquad
d_l = R_S(l) - R_I(l),
\qquad
E = \{l \in T : |S(l)| \geq m,\ |I(l)| \geq m\},
\label{eq:within}
\end{equation}
where $E$ is the set of tail labels estimable in both contexts. We report $d$ with the standard error of the mean over $E$, $\mathrm{se}(d) = \mathrm{sd}\bigl(\{d_l\}_{l \in E}\bigr)/\sqrt{|E|}$. We use $m = 1$, the definedness floor of Definition~\ref{def:partition}. Larger $m$ trades labels for per-label precision and is swept in Appendix~\ref{app:sensitivity}.

$E$ is a subset of $T$, not $T$ itself, and $d$ says nothing about labels outside it. Membership requires positives on both sides of the partition and so depends on $\pi(l)$ and $n_l$: a label that never appears without a companion is excluded, not scored as maximally dependent. Membership is decided by gold counts alone, independently of any model, so conditioning on $E$ leaves each $d_l$ unbiased for its label's contrast. We report $|E|$ beside every $d$, so that ``could not be estimated'' stays a different statement from ``estimated as zero'', and report $d$ as undefined when $|E| < 5$.

Both $R_S$ and $R_I$ are computed from thresholded predictions. We fit one threshold per label on the validation split and freeze it before evaluation, the convention the audited systems are tuned under. A single global threshold would confound the tail's calibration with its competence.

Corpora are compared on $d$ as defined, which is on a common recall scale. Arms on one corpus are compared on the relative form
\begin{equation}
d_{\text{rel}} = \frac{d}{\bar{R}_E},
\qquad
\bar{R}_E = \frac{1}{|E|} \sum_{l \in E} R(l),
\label{eq:relative}
\end{equation}
the effect as a fraction of the model's own recall on the labels it is measured over, because a weak model has little recall for context to redistribute and its absolute effect is small however completely it depends on companions. $d_{\text{rel}}$ can exceed $1$. With $R_I = 0$ on every label it approaches the reciprocal of the recall-weighted test-split support ratio, so a strongly dependent arm on a corpus whose estimable labels have $\pi(l)$ near $1/2$ can read close to $2$.

\subsection{Capability must be co-reported}
\label{sec:rule3}

We call a trained model variant an arm and the variant identical to it but for one mechanism its control. An arm is compared with its control on the same corpus, as Definition~\ref{def:damage} specifies. A dependence measure can always be reduced by damaging a model. Predicting less on supported positives lowers $d$ without anything having been learned about the independent ones. Equation~\ref{eq:within} is therefore read with independent recall beside it.

\begin{definition}[Correction, damage and worsening]
\label{def:damage}
Let $\Delta$ be an arm's change in $d_{\text{rel}}$ against its control, and $\sigma(\cdot)$ that
change in standard errors: for objectives, the mean paired difference with BCE across seeds over
its standard error; for mechanism arms, the unpaired separation of the arm's runs from the pooled
runs of its family's controls, $|\bar{x}_a - \bar{x}_c| / \sqrt{s_a^2/n_a + s_c^2/n_c}$
(Section~\ref{sec:protocol}). A contrast is \emph{flagged} where $|\sigma(\Delta)| > 2$ and is then
read with $\sigma(R_I)$:
\begin{itemize}
\item \emph{Correction}: $\Delta$ falls and $\sigma(R_I) > +2$.
\item \emph{Damage}: $\Delta$ falls and $\sigma(R_I) \leq +2$; the effect fell with nothing gained
  where the measure exists to look.
\item \emph{Worsening}: $\Delta$ rises and $\sigma(R_I) < -2$.
\item \emph{Worsening}$^{*}$: $\Delta$ rises and $\sigma(R_I) \geq -2$.
\end{itemize}
For mechanism arms only falls are read, since a rise is the constructed signature of the positive
controls. A contrast that is not flagged is not called, which on few seeds is a weaker statement
than a null.
\end{definition}

The bar of $|\sigma| > 2$ is a screening threshold, not a significance test. On five seeds it corresponds to $p \approx 0.12$ two-sided. Where we describe a result as significant we mean that it survives a Holm--Bonferroni correction across all tested contrasts, and we complement the seed-level $\sigma$ with a hierarchical bootstrap over labels and seeds (Section~\ref{sec:protocol}). Reading outcomes four ways rather than two is what lets damage be told from correction. Section~\ref{sec:objectives} presents an objective that lowers $d$ more than any other yet fails Definition~\ref{def:damage}, and Section~\ref{sec:derive} a corpus where deriving arms fall below their controls for the same reason.

\subsection{Screening corpora}
\label{sec:screening}

Because $\pi$ is computed from the label matrix alone, a corpus can be assessed before any model is trained. We use four statistics, labeled S1--S4 and referred to by label throughout.

\begin{enumerate}
\item[\textbf{S1}] \textbf{Mean support ratio $\bar{\pi}$.} By Corollary~\ref{cor:screening}, it bounds the inflation of reported tail recall, provided the support ratio transfers to the evaluation split. A corpus with small $\bar{\pi}$ cannot inflate reported tail recall materially, though it may still contain individual labels that are wholly companion-dependent. $\bar{\pi}$ bounds the inflation, it does not predict the size of $d$.
\item[\textbf{S2}] \textbf{Labels per document.} Below roughly $1.5$, most positives carry no other label and the phenomenon has little room to occur. Section~\ref{sec:screening-results} reports the exceptions to this heuristic.
\item[\textbf{S3}] \textbf{Companion richness and positives per tail label.} A mechanism that computes tail scores from companion labels needs companions to compute from. Where mean $|C(l)|$ is low, its measured dependence is diluted by labels with no companion. And a mapping from companion pattern to tail label must be fitted from each label's own positives, so a corpus with few training positives per tail label cannot support such a mechanism however rich its companion sets are. A corpus failing S3 still admits the measure, but a mechanism arm run on it cannot be read as evidence about the mechanism.
\item[\textbf{S4}] \textbf{$|E|$ on the evaluation split.} The label space must be large enough for a head--tail split and a companion relation to form, with enough tail labels carrying positives in both contexts. Where $|E| < 5$, the estimand is undefined.
\end{enumerate}

If the audit proceeds, $d$ is reported with $|E|$ beside it and read under the co-report of Definition~\ref{def:damage}. Section~\ref{sec:screening-results} tests S1--S4 against the corpora we audited.

\section{Experimental setup}
\label{sec:setup}

We hold the encoder fixed within each corpus and re-implement every candidate method within a shared training framework, so that a comparison within a corpus is a comparison of objectives or mechanisms and not of backbones or capacity. This section describes the datasets, the models and objectives, the mechanism arms, the protocol, and one arm that is not a trained classifier.

\subsection{Datasets}
\label{sec:datasets}

We screen \nScreened{} text corpora in \nClusters{} independent clusters, spanning EU law~\citep{chalkidisLargeScaleMultiLabelText2019,lozamenciaEfficientPairwiseMultilabel2008,chalkidisLexGLUEBenchmarkDataset2022a}, biomedical indexing~\citep{hershOHSUMEDInteractiveRetrieval1994,tsatsaronisOverviewBIOASQLargescale2015}, a retail taxonomy~\citep{mcauleyImagebasedRecommendationsStyles2015}, Reuters newswire~\citep{lewisRCV1NewBenchmark}, academic abstracts~\citep{yangSGMSequenceGeneration2018}, Wikipedia tags~\citep{zubiagaEnhancingNavigationWikipedia2009} and Chilean (Spanish) consumer contracts~\citep{lofflerPredictingPotentiallyAbusive2025,loefflerRetrievalAugmentedDetectionPotentially2026}. Several corpora are variants of one another (three truncations of one EUROVOC vocabulary, two corpora sharing a MeSH-derived thesaurus, four views over one set of clauses), so we report cluster counts wherever an analysis would otherwise be pseudo-replicated. Table~\ref{tab:register} in Appendix~\ref{app:register} lists every screened corpus. We exclude vision corpora deliberately.

Table~\ref{tab:datasets} describes the corpora we audit, those with enough estimable tail labels to carry the estimand, with one worked tail label each. We use each corpus's published splits. Ohsumed's distributed test split is larger than its training split (293,855 against 54,709 documents), which we keep as published. Corpora without a validation split have one carved from their training split, and every decision threshold is fitted on it.

\begin{table*}[pos=t]
\footnotesize
\setlength{\tabcolsep}{3pt}
\caption{The audited corpora. Splits are in documents; Ohsumed, PubMed MeSH, Wiki10-31K and AmazonCat-13K publish no validation split, so one is carved from training (Section~\ref{sec:protocol}). $|L|$ is the label count, card.\ the mean labels per training document, $|T|$ the tail at the reported split, tokens the mean whitespace tokens per test document. The example is the tail label with the most test positives among those estimable in both contexts, with its companions $C(l)$ from the training labels, strongest first, and its test counts $|S(l)|/|I(l)|$. EUROVOC concepts are shown by their English preferred labels; Wiki10-31K and AmazonCat-13K by their distributed tag and category names.}
\label{tab:datasets}
\begin{tabularx}{\textwidth}{@{}lp{3.2cm}rrrrrrrX@{}}
\toprule
Corpus & Domain; vocabulary & train & val & test & $|L|$ & card. & $|T|$ & tokens & Example tail label: companions ($|S|/|I|$) \\
\midrule
EURLEX57K & EU legislation; EUROVOC thesaurus & 45{,}000 & 6{,}000 & 6{,}000 & 4{,}271 & 5.07 & 2{,}136 & 350 & \emph{Friuli-Venezia Giulia}: State aid (1/4) \\
PubMed MeSH & PubMed abstracts; MeSH & 40{,}000 & -- & 10{,}000 & 21{,}918 & 12.69 & 10{,}959 & 206 & \emph{Galactosyltransferases}: Animals; RNA, Messenger (5/3) \\
Ohsumed & MEDLINE abstracts; MeSH & 54{,}709 & -- & 293{,}855 & 14{,}623 & 10.32 & 7{,}312 & 172 & \emph{knowledge, attitudes, practice}: adult; family planning; infant, newborn; sex behavior (85/74) \\
Wiki10-31K & Wikipedia articles; social tags & 14{,}146 & -- & 6{,}616 & 30{,}938 & 18.64 & 15{,}469 & 2052 & \emph{exercises}: dynamics; e; education; effective; +44 more (4/1) \\
AmazonCat-13K & product descriptions; retail taxonomy & 1{,}186{,}239 & -- & 306{,}782 & 13{,}330 & 5.04 & 6{,}665 & 210 & \emph{ages 9-12}: by age; children's; magazine subscriptions; entertainment (15/1) \\
LexGLUE EURLEX & EU legislation; 100 EUROVOC concepts & 55{,}000 & 5{,}000 & 5{,}000 & 100 & 4.42 & 50 & 1871 & \emph{2441 budget}: 1021 EU finance (96/95) \\
\bottomrule
\end{tabularx}
\end{table*}

\subsection{Models and objectives}
\label{sec:models}

Every loss-side run is a label-wise attention classifier~\citep{mullenbachExplainablePredictionMedical2018} on a pretrained transformer encoder chosen per corpus for its domain, i.e., Legal-BERT~\citep{chalkidisLEGALBERTMuppetsStraight2020} on the two EUROVOC corpora, BiomedBERT~\citep{guDomainSpecificLanguageModel2022} on the two MeSH corpora, and BERT-base~\citep{devlinBERTPretrainingDeep2019} on Wiki10-31K, AmazonCat-13K, AAPD and arXiv-157, for which no in-domain encoder is an obvious fit. We train with AdamW~\citep{loshchilovDecoupledWeightDecay2019} at a learning rate of $3 \times 10^{-5}$ for the encoder and $10^{-3}$ for every other parameter, batch size 16, linear warm-up over the first 10\% of steps, weight decay 0.01, gradient clipping at 1.0, and early stopping on validation loss at a patience of two epochs. Table~\ref{tab:setup} gives, per corpus, the encoder, input length, training budget and number of objectives, and the tail micro-F1 of the binary cross-entropy baseline, which lies between $0.058$ and $0.168$ on the five corpora whose label spaces run into the thousands and between $0.428$ and $0.525$ on the three with at most 130 labels.

We evaluate \nObjectives{} loss-side objectives: binary cross-entropy (BCE), focal~\citep{linFocalLossDense2018}, asymmetric~\citep{ben-baruchAsymmetricLossMultiLabel2021}, class-balanced~\citep{cuiClassBalancedLossBased2019}, distribution-balanced~\citep{wuDistributionBalancedLossMultiLabel2021}, logit adjustment~\citep{menonLongtailLearningLogit2021a} and LDAM~\citep{caoLearningImbalancedDatasets2019}. The full grid of objectives is run at five seeds on \nFullGrid{} corpora. Of these, AAPD carries no estimable effect ($|E| < 5$), so objectives are compared on the remaining three. Recent long-tailed multi-label objectives designed for other tasks, dynamic loss balancing for document-level relation extraction~\citep{zhangDynamicLossBalancing2026} and label-knowledge distillation for clinical coding~\citep{caiLabelsMatterIncorporating2026}, are outside the loss-side family we audit and are not evaluated.

\subsection{Mechanism arms}
\label{sec:arms}

Correlation-exploiting methods are evaluated as ladders: a family's arms, each compared with a control from the same architecture with the mechanism disabled, holding capacity and training fixed where the family allows it. We re-implement four families from the literature and one of our own under the common protocol. Table~\ref{tab:arms} specifies every arm. No arm reads gold labels at inference. Gold labels enter only as co-occurrence statistics estimated on the training split, and, for one arm, as teacher-forced inputs during training.

We sort the arms by where the tail label's score comes from. An adjusting arm keeps a tail score computed from the document and modifies it with label correlation. We count the graph family in the manner of \citet{chenMultiLabelImageRecognition2019} (with GATv2 rather than GAT~\citep{velickovicGraphAttentionNetworks2018}, since GAT's attention is static~\citep{brodyHowAttentiveAre2022}), the correlation network of \citet{xunCorrelationNetworksExtreme2020}, and an inference-time co-occurrence reranker in the manner of \citet{yanLabelCoRankRevolutionizingLong2025a} as adjusting mechanisms. A deriving arm computes the tail score from other labels' predictions. Here, we count the sequence decoder of \citet{yangSGMSequenceGeneration2018}, which conditions each step on the labels it has already emitted, its label-only variant, whose tail scores see no document and only the label sequence the decoder generates, and the substitute arm of a family we construct so that its two arms differ only in whether a tail label's score is replaced by evidence from the top predicted labels (substitute) or adjusted by it (gate). The label-only and substitute arms are constructed positive controls in the spirit of partial-input baselines~\citep{gururanganAnnotationArtifactsNatural2018,poliakHypothesisOnlyBaselines2018}. Nothing is withheld from the input, but where the tail score may be computed from is constrained. The sequence-decoding family pools the encoder output rather than using label-wise attention, so its control is the plain BCE baseline and its rows carry a capacity difference the other families do not.

\begin{table*}[t]
\footnotesize
\setlength{\tabcolsep}{4pt}
\caption{Mechanism arms. Every arm shares the corpus's encoder, optimizer, BCE objective and per-label validation thresholds. No arm reads gold labels at inference. ``pred.'' denotes the model's own scores, and co-occurrence statistics are estimated on the training split. $z_l$ is the baseline logit, $r_l$ the label-wise attention vector, $A$ the row-normalized training co-occurrence graph ($\tau = 0.4$, smoothing $0.25$), $\delta_l$ a learned correction from label-query attention over the document's tokens, $e_l$ a linear readout of the mean embedding of the top five predicted labels, and $h_t$ the decoder state after $t$ emitted labels. $^{c}$ marks a control.}
\label{tab:arms}
\begin{tabularx}{\textwidth}{@{} l l >{\hsize=1.1\hsize}X >{\hsize=1.3\hsize}X >{\hsize=0.6\hsize}X @{}}
\toprule
Family & Arm & Input at inference & Tail score & Control \\
\midrule
Graph (adjusting) & identity$^{c}$ & document & $w_l \cdot r_l$, $w_l$ from a label embedding, $A = I$ & -- \\
 & co-occurrence & document & $w_l$ from a two-layer GCN over $A$ & identity \\
 & GATv2 & document & as above, second layer attention over $A$'s edges & identity \\
 & dynamic & document, companions' attention vectors & GCN $w_l \cdot (A r)_l$ & identity \\
\addlinespace
Correlation network (adjusting) & diagonal$^{c}$ & own pred. & $z_l + a_l \sigma(z_l) + c_l$ & -- \\
 & frozen$^{c}$ & all pred. & $z_l + U\,\mathrm{elu}(D\sigma(z))$, $D$ random and frozen & -- \\
 & CorNet & all pred. & $z_l + U\,\mathrm{elu}(D\sigma(z))$, two learned blocks & diagonal, frozen \\
\addlinespace
Reranking (adjusting) & seed-only$^{c}$ & top-10 pred., document & $z_l + \delta_l$ on the ten seed labels & -- \\
 & random$^{c}$ & top-10 pred., random table & $z_l + \delta_l$ on seeds and ten random promotions & -- \\
 & co-occurrence & top-10 pred., training $P(j \mid l)$ & $z_l + \delta_l$ on seeds and ten companion promotions & seed-only, random \\
\addlinespace
Sequence (deriving) & BCE$^{c}$ & document & $z_l$ (label-wise attention) & -- \\
 & SGM & pooled document, own emitted labels & $\max_t (q + h_t) \cdot o_l$, greedy, ten steps & BCE \\
 & SGM, teacher-forced & as SGM; gold label prefix in training & as SGM & BCE \\
 & label-only & label chain from the first emitted label & $\max_{t \geq 1} h_t \cdot o_l$, no document & BCE \\
\addlinespace
Derivation & gate$^{c}$ (adjusting) & document, top-5 pred. & $z_l + \beta \tanh(e_l)$ & -- \\
 & substitute (deriving) & top-5 pred. & $e_l$, document score discarded & gate \\
\bottomrule
\end{tabularx}
\end{table*}

\subsection{Protocol}
\label{sec:protocol}

Loss-side objectives are run at five seeds and paired with BCE by seed. Ladder arms are reported where they have at least three runs. The family's controls are pooled, run-weighted, into one baseline, and an arm's $\sigma$ is its separation from that baseline in standard errors. A contrast is flagged at $|\sigma| > 2$ (Definition~\ref{def:damage}). Significance claims use a Holm--Bonferroni correction across all objective--corpus cells. Because seed variance ignores label sampling, which dominates on corpora with few estimable labels, objective contrasts are also given a hierarchical bootstrap. Each of 2,000 draws resamples the estimable labels with replacement, shared across the runs of a corpus, and the paired seeds with replacement, and recomputes the mean paired difference in $d_{\text{rel}}$.

We report at $\tau = 0.5$ with the tail as the bottom half of labels by training frequency and $m = 1$, with $\lambda = c = 1$, and sweep $\tau$, the head--tail split and $m$ in Appendix~\ref{app:sensitivity}. Decision thresholds are fitted per label on the validation split and frozen before evaluation.

\subsection{A system that is not a classifier}
\label{sec:rag-setup}

To show that the evaluator needs only prediction files, we also audit one retrieval-augmented language model on Ohsumed. Documents and label names are embedded with the 33M-parameter all-MiniLM sentence encoder~\citep{wangMiniLMDeepSelfAttention2020,reimersSentenceBERTSentenceEmbeddings2019b}, served as \texttt{all-minilm:33m}, the 40 labels whose names are most similar to a document are retrieved, and Gemma~4 E2B~\citep{teamGemma4Technical2026} selects among them by name at temperature zero, with the prompt given in Appendix~\ref{app:prompt}. The arm produces label sets without logits and is evaluated by the same code as every other arm. Retrieval reads documents and label names and never label co-occurrence, so it cannot learn companion structure from the labels. However, this does not mean that it is companion-neutral, because a document that carries a tail label's companions may be more topical for that label's name, which is what Section~\ref{sec:rag} measures.

\section{Results}
\label{sec:results}

We first establish that the estimand measures companion dependence and that the effect is present (RQ1), then ask whether objectives move it (RQ2), how much of it training labels reveal (RQ3), and which design property moves it (RQ4). Finally we apply the evaluator, unchanged, to a system that is not a classifier.

\subsection{Validity of the instrument (RQ1)}
\label{sec:validity}

Before measuring companion dependence, we check that the estimand measures it, i.e., whether it improves on the constructions of Section~\ref{sec:instrument}, and whether it responds when dependence is built into a model.

{\floatplacement{table}{H}
\begin{table}[t]
\caption{Independence conditioning by corpus. $\pi_{\text{pool}}$ is the support ratio pooled over tail occurrences on the test split. The middle group is the within-label estimand. $d$ is the effect of Equation~\ref{eq:within} for binary cross-entropy, the headline estimate, averaged over the BCE runs, with a normal 95\% interval from the standard error over labels. $d_{\text{all}}$ pools every loss-side objective run on the corpus and is given as a robustness check, since objectives move $d$ (Section~\ref{sec:objectives}). $|E|$ is the number of tail labels estimable in both contexts. A dash marks an effect that is undefined because $|E| < 5$, which is not the same as an effect of zero. The right group is the difference-of-F1 construction of Equation~\ref{eq:diff-f1}, the confounded quantity and its confound. Gap is the share of reported tail F1 that it attributes to conditioning. Null is the share of that gap which a base-rate null already explains, using a conservative per-label null over five BCE seeds (Appendix~\ref{app:null}).}
\label{tab:corpus}
\begin{tabular*}{\tblwidth}{@{\extracolsep{\fill}}lr@{\hspace{1.6em}}rrrr@{\hspace{1.6em}}rr@{}}
\toprule
 & & \multicolumn{4}{c}{Within-label estimand} & \multicolumn{2}{c}{Difference of F1} \\
\cmidrule(lr){3-6} \cmidrule(l){7-8}
Corpus & $\pi_{\text{pool}}$ & $d$ & 95\% CI & $d_{\text{all}}$ & $|E|$ & gap & null \\
\midrule
EURLEX57K & 0.262 & $+0.280$ & $\pm0.199$ & $+0.322$ & 35 & $-24.6$\% & $141$\% \\
PubMed MeSH & 0.401 & $+0.138$ & $\pm0.055$ & $+0.138$ & 357 & $-38.0$\% & $96$\% \\
Wiki10-31K & 0.241 & $+0.133$ & $\pm0.049$ & $+0.124$ & 260 & $+14.3$\% & $-67$\% \\
Ohsumed & 0.147 & $+0.131$ & $\pm0.014$ & $+0.128$ & 1{,}986 & $-49.3$\% & $103$\% \\
AmazonCat-13K & 0.281 & $+0.084$ & $\pm0.029$ & $+0.084$ & 910 & $-17.2$\% & $123$\% \\
LexGLUE EURLEX & 0.080 & $+0.073$ & $\pm0.115$ & $+0.073$ & 7 & $-1.6$\% & $137$\% \\
AAPD & 0.084 & -- & -- & -- & 4 & $+2.3$\% & $-197$\% \\
arXiv-157 & 0.018 & -- & -- & -- & 2 & $+2.2$\% & $-63$\% \\
\bottomrule
\end{tabular*}
\end{table}

If the between-label constructions of Section~\ref{sec:instrument} are confounded, they should disagree with the within-label estimand, and the disagreement should be explained by the confound rather than by companion dependence.
Table~\ref{tab:corpus} reports, for every audited corpus, the estimand $d$ beside the difference-of-F1 gap of Equation~\ref{eq:diff-f1}. The null column gives the share of that gap a base-rate null already explains, i.e., a model that keeps each label's own recall and false-positive rate and depends on context in no other way (Appendix~\ref{app:null}).
The null explains the entire gap, or more, on five of the six estimable corpora, from 103\% on Ohsumed to 141\% on EURLEX57K. On Wiki10-31K the gap has the opposite sign. It reads $+14.3\%$, reporting that conditioning raises tail F1, while $d$ reads $\WikiEffectBce$ over \WikiLabels{} labels. The pooled-ratio construction fails in its own way, reading $\OhsumedPooledRatio$ on Ohsumed against $\OhsumedWithinRatio$ within labels.
The obvious constructions are dominated by a base-rate artifact and can reverse the sign of the phenomenon. We therefore report the within-label $d$ throughout and do not report the gap as a finding.

An arm that depends on companions by construction should read far above its own control.
The label-only arm of the sequence-decoding family computes tail scores from no document at all, see Table~\ref{tab:arms}. We compare its relative effect (Equation~\ref{eq:relative}) with the BCE baseline.
In Table~\ref{tab:ladders} the label-only arm reads $\LadOhsumedBlind$ against the baseline's $\LadOhsumedBase$ on Ohsumed ($\LadOhsumedBlindSigma$ standard errors) and $\LadEurlexBlind$ against $\LadEurlexBase$ on EURLEX57K ($\LadEurlexBlindSigma$ standard errors). Its absolute effect is small, because an arm built this way predicts little, which is why arms are compared on the relative form. A relative value above 1 is expected for such an arm (Section~\ref{sec:estimand}).
The estimand responds to dependence we know is present, in the expected direction and by a wide margin.

We also sought the converse check, that a genuine improvement on a corpus without companion structure survives conditioning. The grid offers no case to test it on. The one tail-F1 improvement that clears twice its seed noise, class-balanced loss on AAPD ($+0.021$ in tail F1), is a precision gain. On recall it reads $\SurvRecallDelta$ at $\SurvRecallSnr$ standard errors, so there is no recall improvement for a recall-based estimand to condition. We leave the check untested rather than report a share of F1, which would rest on the construction rejected above.

\subsection{Presence and size of the effect (RQ1)}
\label{sec:establish}

If a model borrows tail recall from companions, it should recover a tail label more often in documents that carry one of its companions than in documents that do not, so $d$ should be positive on corpora where companions exist.
We compute $d$ by Equation~\ref{eq:within} for the BCE baseline on every audited corpus, with a 95\% interval from the standard error across estimable labels, and give the mean over every loss-side run as a robustness column, since objectives move $d$ (Section~\ref{sec:objectives}).
Table~\ref{tab:corpus} shows that $d$ is positive on all \nEstimableCorpora{} corpora where it is defined, and its interval excludes zero on all but one. The most precise estimate is Ohsumed's, $\OhsumedEffectBce$ over \OhsumedLabels{} labels ($\pm\OhsumedEffectBceCI$). The largest is EURLEX57K's, $\EurlexEffectBce$, which rests on \EurlexLabels{} labels with an interval of $\pm\EurlexEffectBceCI$. The exception is LexGLUE EURLEX, whose seven estimable labels give an interval that includes zero. On AAPD and arXiv-157 $|E|$ falls below five and the effect is undefined. These are the two corpora with the least companion structure ($\pi_{\text{pool}}$ of $0.084$ and $0.018$). The pooled column agrees in sign everywhere. Appendix~\ref{app:sensitivity} shows that the sign holds in every cell of the $(\tau, \text{tail})$ grid and at every floor $m$ at which the effect remains estimable.
Companion dependence is present wherever it can be measured. Its magnitude varies by a factor of about four across corpora, and $d$ and $|E|$ are read together.

A remaining explanation is that independent occurrences sit in harder documents, for instance shorter ones, which would lower $R_I$ without any shortcut being involved.
We match supported and independent occurrences on document length, in whitespace tokens and within each label, and recompute the recall difference $R_S - R_I$ on the matched set for one BCE seed.
Matching moves the difference from $+0.367$ to $+0.365$ on Ohsumed (29,817 matched occurrences), from $+0.354$ to $+0.356$ on EURLEX57K, and leaves it unchanged at three decimals on PubMed MeSH. The check pools the recall difference over occurrences, because matching operates on occurrences, so it tests whether length drives the contrast rather than the size of $d$. EURLEX57K contributes only 216 matched occurrences, so Ohsumed is the informative case.
Document length explains essentially none of the effect.

\subsection{Objectives do not move the effect consistently (RQ2)}
\label{sec:objectives}

\begin{table}[t]
\caption{Every loss-side objective against binary cross-entropy, on each corpus carrying a full grid and an estimable effect. $\Delta$ is the change in the relative within-label effect and $\sigma$ its paired $t$ across five seeds, i.e., the separation of the arm from binary cross-entropy in standard errors on 4 degrees of freedom, where $|\sigma| = 2$ is $p \approx 0.12$ two-sided and $|\sigma| = 4$ is $p \approx 0.016$. A cell is flagged at $|\sigma| > 2$, a screening bar and not a significance test. Holm--Bonferroni across all 18 tested cells leaves 1 of the 5 flagged cells significant at $0.05$. It drops class-balanced on EURLEX57K, distribution-balanced on Wiki10-31K, focal on Ohsumed and logit adjustment on Ohsumed, each of which we mark with a dagger. Table~\ref{tab:objectives-boot} gives the 95\% interval of $\Delta$ from a hierarchical bootstrap over labels and seeds. ``Worsens'' marks a rise in the effect with independent recall falling, and ``damage'' a fall obtained without independent recall rising (Section~\ref{sec:rule3}). An asterisk marks a rise whose independent-recall term does not clear the bar. No objective reads the same way on two corpora.}
\label{tab:objectives}
\begin{tabular*}{\tblwidth}{@{\extracolsep{\fill}}l@{\hspace{1.6em}}rr@{\hspace{1.6em}}rr@{\hspace{1.6em}}rr@{}}
\toprule
 & \multicolumn{2}{c}{Ohsumed} & \multicolumn{2}{c}{EURLEX57K} & \multicolumn{2}{c}{Wiki10-31K} \\
\cmidrule(lr){2-3} \cmidrule(lr){4-5} \cmidrule(l){6-7}
Objective & $\Delta$ & $\sigma$ & $\Delta$ & $\sigma$ & $\Delta$ & $\sigma$ \\
\midrule
focal & $+0.037$\,\textsuperscript{w}\textsuperscript{\dag} & $+4.7$ & $+0.067$ & $+0.7$ & $+0.025$ & $+0.5$ \\
asymmetric & $+0.000$ & $+0.0$ & $+0.060$ & $+0.7$ & $-0.038$ & $-1.0$ \\
class-balanced & $+0.019$ & $+1.7$ & $+0.204$\,\textsuperscript{w}\textsuperscript{\dag} & $+4.0$ & $+0.055$ & $+0.9$ \\
distribution-balanced & $+0.043$ & $+1.3$ & $+0.244$\,\textsuperscript{w} & $+7.5$ & $-0.156$\,\textsuperscript{d}\textsuperscript{\dag} & $-4.0$ \\
logit adjustment & $+0.026$\,\textsuperscript{w*}\textsuperscript{\dag} & $+2.5$ & $+0.106$ & $+1.3$ & $-0.018$ & $-0.4$ \\
LDAM & $-0.002$ & $-0.3$ & $+0.058$ & $+1.0$ & $+0.011$ & $+0.6$ \\
\bottomrule
\end{tabular*}
\par\smallskip{\footnotesize\textsuperscript{w} worsens, \textsuperscript{w*} a rise without independent recall falling at the bar, \textsuperscript{d} damage, \textsuperscript{c} corrects, \textsuperscript{\dag} flagged but does not survive Holm.}
\end{table}

If a long-tail objective addressed companion dependence, it should lower the relative effect while raising independent recall, a correction by Definition~\ref{def:damage}, and do so on more than one corpus.
We pair each of six objectives with BCE by seed on the three corpora that carry a full grid and an estimable effect, \nObjectiveCells{} cells in total, flag each at $|\sigma| > 2$, apply a Holm correction across all \nObjectiveCells{}, and give each cell the hierarchical bootstrap of Section~\ref{sec:protocol} (Table~\ref{tab:objectives-boot}).
Table~\ref{tab:objectives} contains no correction. \nObjectiveCalled{} of the \nObjectiveTested{} cells are flagged. Four are rises: focal loss on Ohsumed by \FocalDRel{} at \FocalTRel{} standard errors, while independent recall falls at \FocalTRI{}, class-balanced and distribution-balanced losses on EURLEX57K, and logit adjustment on Ohsumed, whose independent-recall term does not clear the bar. The fifth is damage: distribution-balanced loss lowers the effect on Wiki10-31K by \DistBalDRel{} at \DistBalTRel{} standard errors, the largest reduction in the grid, while independent recall does not move (\DistBalTRI{} standard errors) and supported recall falls (\DistBalTRS{}). Without the co-report this cell would read as a correction. After Holm correction \nObjectiveHolm{} of the \nObjectiveCalled{} flagged cells is significant at $0.05$, distribution-balanced loss on EURLEX57K. The bootstrap, which also resamples labels, is stricter still where labels are few. \BootExclude{} of \BootCells{} intervals exclude zero, \BootFlaggedExclude{} of the \BootFlagged{} flagged cells, focal loss and logit adjustment on Ohsumed, both rises, while the EURLEX57K cells rest on \EurlexLabels{} labels and their intervals include zero. No objective is flagged on more than one corpus, and distribution-balanced loss is flagged in opposite directions on two.
No objective is shown to reduce companion dependence, two are shown to raise it on Ohsumed, and none behaves the same way on a second corpus. An objective's effect on companion dependence is a property of the objective and the corpus together.

Asking whether a tail improvement survives conditioning presupposes that there is an improvement to condition.
We pair each objective with BCE by seed on tail F1 and compare each difference with its own seed noise.
On EURLEX57K none of the six contrasts reaches twice its seed noise. The largest signal-to-noise ratio is $1.58$, and the tail-F1 differences run from $-0.005$ to $+0.005$. On Ohsumed four of the six do, and all four are losses. Focal, asymmetric, distribution-balanced and logit-adjusted losses each lower tail F1 against BCE, by $0.021$ to $0.026$. Across the grid, the AAPD contrast of Section~\ref{sec:validity} is the only improvement that clears the floor.
On the corpora where companion structure exists, no objective improves tail F1 on BCE beyond its seed noise, and on Ohsumed most are measurably worse. The variation that matters lies between corpora, which raises the question of how much of it the corpus reveals before training.

\subsection{What the corpus reveals before training (RQ3)}
\label{sec:screening-results}

\begin{figure}
\centering
\includegraphics[width=0.72\linewidth]{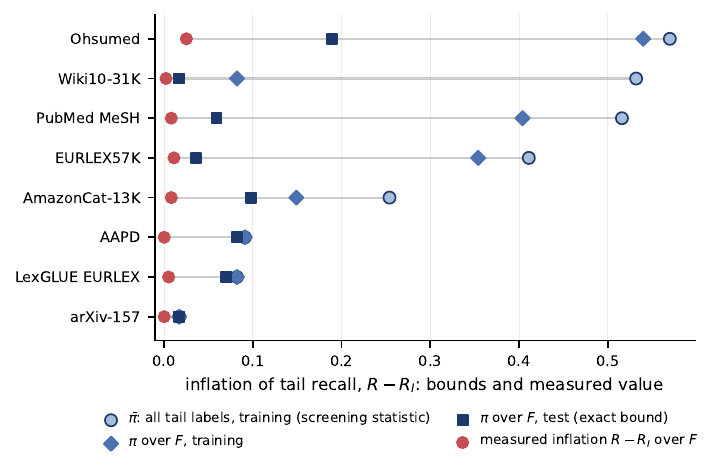}
\caption{Where the slack of the screening bound goes. Each row is an audited corpus, ordered by its screening statistic, and reads from right to left. The light circle is $\bar{\pi}$, computed from training labels over the whole tail. The diamond is the support ratio over $F$, the tail labels whose independent recall is defined on the test split, still on training labels. The square is the same ratio on the test split, which bounds the inflation exactly (Corollary~\ref{cor:screening}). The red dot is the measured tail-mean inflation of reported recall, $R - R_I$ over $F$, for the BCE runs. Each step removes one source of slack: restricting the tail to $F$, moving from training to test labels, and the model itself. Values are in Table~\ref{tab:inflation}.}
\label{fig:screening}
\end{figure}

\begin{table}[t]
\caption{The screening bound against the quantity it bounds. $F$ is the set of tail labels whose independent recall $R_I$ is defined on the test split. Inflation is the mean of $R(l) - R_I(l)$ over $F$, averaged over the binary cross-entropy runs. $\pi_{\text{test}}$ over $F$ bounds it exactly (Corollary~\ref{cor:screening}, with $C(l)$ fixed from training); $\pi_{\text{train}}$ over $F$ and $\bar{\pi}$, the screening statistic over the whole tail, are available before training. The last column is the share of tail labels with $\pi_{\text{train}}(l) \geq 0.9$, on which the per-label bound is close to vacuous.}
\label{tab:inflation}
\begin{tabular*}{\tblwidth}{@{\extracolsep{\fill}}l@{\hspace{1.6em}}rr@{\hspace{1.6em}}r@{\hspace{1.6em}}rrr@{}}
\toprule
 & \multicolumn{2}{c}{Measured on test} & \multicolumn{1}{c}{Exact bound} & \multicolumn{3}{c}{Known before training} \\
\cmidrule(lr){2-3} \cmidrule(lr){4-4} \cmidrule(l){5-7}
Corpus & $|F|$ & inflation & $\pi_{\text{test}}$ on $F$ & $\pi_{\text{train}}$ on $F$ & $\bar{\pi}$ & $\pi \geq 0.9$ \\
\midrule
EURLEX57K & 470 & $0.011$ & 0.036 & 0.354 & 0.411 & 23\% \\
PubMed MeSH & 3{,}133 & $0.008$ & 0.059 & 0.404 & 0.516 & 44\% \\
Ohsumed & 6{,}732 & $0.025$ & 0.189 & 0.540 & 0.570 & 47\% \\
Wiki10-31K & 8{,}342 & $0.002$ & 0.017 & 0.082 & 0.532 & 53\% \\
AmazonCat-13K & 5{,}640 & $0.008$ & 0.098 & 0.149 & 0.254 & 11\% \\
LexGLUE EURLEX & 50 & $0.005$ & 0.070 & 0.082 & 0.082 & 0\% \\
AAPD & 27 & $-0.000$ & 0.082 & 0.091 & 0.091 & 0\% \\
arXiv-157 & 65 & $0.000$ & 0.017 & 0.017 & 0.017 & 0\% \\
\bottomrule
\end{tabular*}
\end{table}

If the screening statistic does what Corollary~\ref{cor:screening} says, the inflation of reported recall measured on test should lie below the exact test-split bound, and the training-label statistic $\bar{\pi}$ should bound it insofar as the support ratio transfers.
For every audited corpus we measure the tail-mean inflation of the BCE runs over the labels whose independent recall is defined, and set it against the test-split support ratio over those labels and against $\bar{\pi}$ (Table~\ref{tab:inflation}, Figure~\ref{fig:screening}).
The inflation lies below both bounds on all \InflNCorpora{} corpora, and far below them. It is at most \InflOhsumed{} (Ohsumed), against a test-split bound of \InflPiTestOhsumed{} and $\bar{\pi} = \InflPiBarOhsumed$. Two things make the training-label bound loose. The support ratio does not transfer. Over the same labels, it is several times lower on the test split than on training on every large-vocabulary corpus, because companion sets estimated on training labels transfer only partly. And the average includes many labels with no supported test positive, whose inflation is zero by construction. A substantial share of tail labels sits at $\pi(l) \geq 0.9$, where the per-label bound is close to vacuous, with \InflVacuousOhsumed\% on Ohsumed and \InflVacuousWiki\% on Wiki10-31K.
The bound is a pre-training ceiling, not a prediction: $\bar{\pi}$ caps reported tail-recall inflation before any model is trained, and on all \InflNCorpora{} corpora the measured value stays below it. The measured value is small for a structural reason: per label $R - R_I = \pi(l)(R_S - R_I) \leq \pi(l) R_S$, so inflation is capped by tail recall, which is low here. Relative to that recall it reaches \InflRelOhsumed{} on Ohsumed, where \InflExcessOhsumed\% of the tail true positives macro-F1 credits are excess (Table~\ref{tab:inflation-scale}), and on \InflOutFSharePubmed\% of PubMed MeSH's tail labels with test positives, reported recall cannot be checked for independence at all. Averaged over $F$, the inflation is close to the product of the test-split support ratio and the within-label effect: on Ohsumed, $\pi_{\text{test}} = \InflPiTestOhsumed$ over $F$ and $d = \OhsumedEffectBce$ multiply to about the measured \InflOhsumed{}, and the product matches Table~\ref{tab:inflation} on every large-vocabulary corpus. The dependence is therefore not small. A companion is present on few test positives of the labels in $F$, because labels whose test positives all carry one fall outside $F$, most tail labels have only one or two test positives, and companion sets transfer only partly from training to test. Over $E$, where companions are present, the inflation reaches \InflEAmazon{} to \InflEEurlex{} on the large-vocabulary corpora (Table~\ref{tab:inflation-scale}). A high $\bar{\pi}$ does not imply high inflation, and it does not predict the size of $d$.

Statistics S2--S4 should predict where the estimand, or a mechanism arm, cannot be read.
We compare the verdict of each, computed from training labels, with what the audit found.
S4 excludes the Chilean abusive-clause corpus, because of its \ChileanLabels{} labels, a split at the median leaves \ChileanTail{} tail labels, of which \ChileanEstimable{} are estimable in both contexts and \ChileanNoCompanion{} have no supported positive at all, although the corpus has over a thousand training documents. S2 separates two corpora from one family, that is, arXiv-157 carries $2.4$ times AAPD's label space but screens five times lower ($0.017$ against $0.091$), because it averages $1.37$ labels per document against $2.41$. This heuristic has exceptions. RCV1-V2 regions ($1.28$ labels per document) and the Chilean dark-pattern corpus ($1.13$) screen at $0.289$ and $0.407$. S3 flags Wiki10-31K, which has the richest companion sets in the register ($|C| = 14.11$) but 30,938 labels against 14,146 training documents, a median of two positives per tail label from which to learn a companion mapping. Section~\ref{sec:derive} shows that its label-only arm collapses due to this. AmazonCat-13K has ample documents but the sparsest companion sets among the corpora on which a mechanism arm is estimable ($|C| = 1.09$), and, as S3 predicts, it carries the smallest derivation effect we observe.
S2--S4 flag each failure we met in the audit, from label statistics alone. One corpus consumed roughly thirty GPU-hours to establish what S3 predicts from a single pass over its labels.

\subsection{Deriving versus adjusting (RQ4)}
\label{sec:derive}

\begin{figure}[pos=t]
\centering
\includegraphics[width=0.6\linewidth]{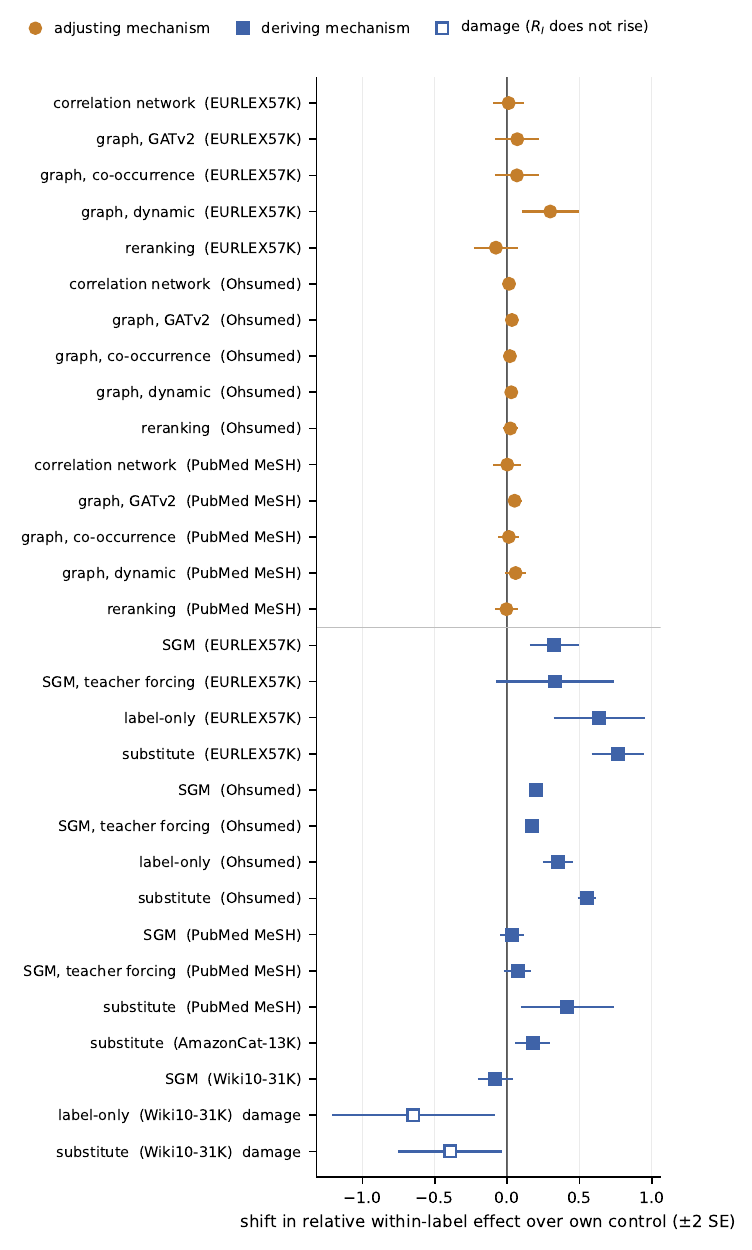}
\caption{Every mechanism-ladder arm measured at three or more runs, as its shift in relative within-label effect over the pooled mean of its own controls, with a bar of two standard errors. A bar clear of zero is a contrast flagged at $|\sigma| > 2$. Adjusting arms (circles) keep a document-based tail score and modify it with label correlation; deriving arms (squares) compute the tail score from other labels' predictions (Table~\ref{tab:arms}). Open markers are damage by Definition~\ref{def:damage}. Table~\ref{tab:ladders} in Appendix~\ref{app:ladders} gives every value.}\label{fig:ladder}
\end{figure}

Since objectives do not move the effect consistently, we test whether it matters if a tail label's score is derived from other labels' predictions or merely adjusted by them.
Each arm of Table~\ref{tab:arms} is compared with its own control on the relative effect.
Figure~\ref{fig:ladder} draws every arm against its control. Fourteen of the fifteen adjusting arms stay within $+0.07$ of their controls. A few on Ohsumed and PubMed MeSH are flagged only because their seeds are tight, and then by at most $+0.05$, and the reranker on EURLEX57K falls slightly below its controls. A pooled control can hide how far its parts disagree. The correlation network's two controls on EURLEX57K read \CornetEurlexControlA{} and \CornetEurlexControlB{}, a spread of \CornetEurlexControlSpread{} against the arm's shift of $\CornetEurlexShift{}$, so that contrast is bounded by its controls rather than read as the mechanism. One adjusting arm does move. The dynamic graph arm on EURLEX57K rises by $+0.30$ at $3.0$ standard errors. However, it does not replicate on Ohsumed ($+0.03$) or PubMed MeSH ($+0.06$, not flagged), although PubMed MeSH's tail receives as large a share of companion evidence through the same fixed graph, and we have no explanation for it. The deriving arms move the effect most. Where flagged, outside Wiki10-31K, they sit clear of zero by $+0.17$ to $+0.76$. On Ohsumed, for instance, SGM moves by $+0.20$ and the label-only arm by $+0.35$, each at more than six standard errors. SGM and its teacher-forced variant on PubMed MeSH ($+0.03$ and $+0.07$), and the teacher-forced variant on EURLEX57K, whose $+0.33$ reaches only $1.6$ standard errors across widely spread seeds, are not flagged. The deriving arms at or below zero are on Wiki10-31K, examined below.
What moves the effect most is how the tail score is built. An arm reads as strongly dependent when its tail score is derived from other labels' predictions, with one unexplained adjusting exception on one corpus.

If the distinction between adjusting and deriving reflects a mechanism, it should replicate across vocabularies and fail only where S3 says the mechanism cannot be learned.
We run the derivation family on three further corpora, namely PubMed MeSH, which shares Ohsumed's MeSH-derived vocabulary, AmazonCat-13K, an independent retail taxonomy, and Wiki10-31K, which fails S3. Every arm is read with the co-report of Definition~\ref{def:damage}.
On PubMed MeSH, substitute reads $\LadPubmedSubst$ against gate at $\LadPubmedGate$ ($\LadPubmedSubstSigma$ standard errors), in the same direction as Ohsumed but noisier. On AmazonCat-13K it reads $\AmazonSubRel$ against $\AmazonGateRel$, a difference of $\AmazonDerivDelta$ at $\AmazonDerivSigma$ standard errors. This is the smallest effect we observe, although the substitute arm there retains more of its control's supported recall (15\% against 4 to 5\% on the MeSH corpora), so capability does not explain the ordering. On Wiki10-31K the deriving arms fall below their controls (substitute $\LadWikiSubst$ against $\LadWikiGate$, the label-only arm $\LadWikiBlind$ against $\LadWikiBase$), but independent recall falls on both, by $0.045$ and $0.052$ at more than ten standard errors, so by Definition~\ref{def:damage} both reductions are damage, drawn as open markers in Figure~\ref{fig:ladder}.
The distinction holds on four corpora in three independent clusters (legal, biomedical and retail vocabularies) and breaks only on the corpus where S3 predicts that a label-only mapping cannot be learned, where the co-report identifies the break as damage rather than counter-evidence. 

\subsection{A system that is not a classifier}
\label{sec:rag}

The evaluator is defined on prediction sets, not on models, so it should apply unchanged to a system that is not a trained classifier and attribute a dependence to the stages of a pipeline.
We audit the retrieval-augmented pipeline of Section~\ref{sec:rag-setup} on Ohsumed with no change to the evaluation code. It emits predictions for \LlmDocuments{} of the \LlmTestDocuments{} test documents (\LlmCoverage\%), so every comparator is restricted to the same documents, on which \LlmLabels{} tail labels are estimable. The arm is a single deterministic run. Because a label outside the shortlist cannot be predicted, we also measure the retriever's ceiling, its recall on supported and on independent tail positives, and the generator's conversion rate within what the retriever reaches.
On the shared documents the pipeline reads $d = \LlmEffect{}$ (relative \LlmRelEffect{}) against BCE's $\LlmBceEffect{}$ (relative \LlmBceRelEffect{}). Part of that dependence belongs to the retriever, which reaches $38.2\%$ of supported tail positives against $33.2\%$ of independent ones although it never reads label co-occurrence. Documents carrying a tail label's companions are apparently more topical for that label's name. Within what it reaches, the generator converts $66.6\%$ against $56.9\%$, so dividing out the ceiling takes the pipeline's ratio of supported to independent recall from $1.34$ to $1.17$. Roughly half of the apparent dependence is retrieval, and a smaller dependence in the generator remains.
The evaluator audits a generative pipeline as readily as a classifier, and the audit is informative only with each stage's ceiling reported beside it. We do not read the lower effect as evidence about language models in general. This is one small model on one corpus and a fraction of its test split, and the generator chooses among 40 candidates where the classifier scores all 14,623 Ohsumed labels.

\section{Discussion}
\label{sec:discussion}

\subsection{Theoretical implications}

Count-based tail measures have a blind spot of quantifiable width. Lemma~\ref{lem:reachable} shows that whenever $\pi(l) \leq R(l) \leq 1 - \pi(l)$ the counts leave the split between supported and independent recall entirely undetermined, so the question of whether a tail label was learned is not one a macro- or micro-averaged score can answer. The answer has to be measured within labels. The two between-label constructions that suggest themselves are confounded, one by a base-rate artifact that reverses its sign on one of our corpora and one by Simpson's paradox. The audit then locates the variation. Companion dependence is present wherever it can be estimated, no loss-side objective reduces it, and what moves it is whether the tail score is derived from other labels' predictions. This places the phenomenon in the corpus and the architecture.

The measurement relates to two existing lines of work. Like worst-group evaluation~\citep{sagawaDistributionallyRobustNeural2020}, it reports the performance an average conceals, but its groups are induced by label co-occurrence and computable from the label matrix, without annotated attributes or a trained model to cluster. Like SCUMBLE~\citep{charteConcurrenceImbalancedLabels2014}, it is computed from labels before training, but it is a proportion of each label's positives tied to a bound on an evaluation measure, and it excludes definitional companions.

\subsection{Practical implications}

The corpora on which we measure the effect are indexed under large annotation vocabularies. MeSH indexes biomedical abstracts, EUROVOC EU legislation, and others on a retail taxonomy and social tags. Automatic indexing with such vocabularies is in production. MEDLINE has been indexed algorithmically since 2022~\citep{medline2022initiativeMEDLINE2022Initiative2021,mtixMTIXNextGenerationAlgorithm2024}; national libraries deploy machine-assisted subject indexing~\citep{poleyAutomaticSubjectCataloguing2025,kasprzikAutomatingSubjectIndexing2023}, and fine-grained semantic indexing below the level of a MeSH descriptor is an active line of work in this journal~\citep{nentidisMeSHFinegrainedSemantic2020}. These systems are evaluated by aggregate precision, recall and F1~\citep{mtixMTIXNextGenerationAlgorithm2024,kritharaRoadManualAutomatic2023}, while rare descriptors remain the hard case~\citep{giustiniAutomatedIndexingMEDLINE2026,poleyAutomaticSubjectCataloguing2025} and audits of indexed records find omissions that affect retrieval~\citep{amar-zifkinAlgorithmicIndexingMEDLINE2025}. An independent occurrence is a document on which a rare descriptor carries information its frequent companions do not, and so a document reachable through that descriptor rather than through the labels it usually accompanies. A classifier that assigns the descriptor only alongside its companions adds little access for a searcher, while its aggregate score reports the index as serviceable. Independent recall measures the access a machine-assigned rare descriptor actually adds, in the sense of evaluating indexing in the context of retrieval~\citep{golubFrameworkEvaluatingAutomatic2016}, and it complements frequency-stratified evaluation~\citep{chalkidisEmpiricalStudyLargeScale2020} by stratifying on co-occurrence rather than on frequency.

For benchmark design, the screening statistics are cheap. $\bar{\pi}$ and S2--S4 cost one pass over a corpus's training labels and say, before any training, whether reported tail recall can be inflated by companions and whether a correlation-exploiting mechanism can be evaluated on the corpus at all.

\subsection{Recommendations for evaluation practice}

The estimand follows a principle of retrieval evaluation under incomplete judgments. Restrict the measure to where it is informative, and report it as undefined elsewhere~\citep{buckleyRetrievalEvaluationIncomplete2004}. Three practices follow.

\emph{Benchmark maintainers} can publish $\bar{\pi}$, with S2--S4, beside a corpus's label statistics. A corpus with a small bound is one on which this failure cannot materially inflate reported tail recall.

\emph{Method authors} can report $d$ beside macro-F1 wherever it is estimable, with its co-report and the label space it was computed in. Importantly, the co-report is not optional. Distribution-balanced loss on Wiki10-31K lowers $d$ more than any other objective and without it would read as a correction (Section~\ref{sec:objectives}). The label space matters because companions are defined against its head set, so per-label effects from different label subsets are not comparable.

\emph{Reviewers} can ask whether a tail result holds on more than one corpus, since no objective in our audit behaves the same way on two. Where a method raises companion dependence, as two objectives do on Ohsumed, the companion-rich corpora show no compensating gain in tail F1 (Section~\ref{sec:objectives}).

None of this requires our estimand. The screening statistic needs only a label matrix, Observation~\ref{prop:invariance} holds for every measure computed from per-label confusion counts, and the evaluator reads prediction files from any producer.

\subsection{Limitations}
\label{sec:limitations}

The scope of the measurement is set by S1--S4, which concern the size and co-occurrence structure of the label space. How the labels were authored makes no difference. An uncontrolled social tag set carries one of the larger effects we measure, and the negative corpora lack scale rather than curation.

Restricted to independent occurrences, hard subsetting is undefined for almost every tail label on the test splits we examined, and reweighting does not recover the missing information. The natural reweighting equals hard subsetting, and the variants that differ reach a bias of $\EstimatorBiasStrongWorst$ against synthetic ground truth where companion strength is at least $\EstimatorStrengthFloor$. The within-label estimand therefore answers only for labels estimable in both contexts and declines to answer for the rest. The screening bound is conservative, by an order of magnitude on our corpora, because the support ratio transfers only partly from training to test.

We propose no remedy. Upweighting independent tail positives is inert. Across $\alpha$ from $0.5$ to $10$, at \CorrectiveMinSeeds{} seeds on EURLEX57K and Ohsumed, the largest rise in independent recall is \CorrectiveMaxDeltaRI{}, at \CorrectiveMaxDeltaRISigma{} standard errors, because those positives are exactly what the phenomenon makes scarce. Penalizing correlation in weight space lowers $d$ on two corpora, but only as damage by Definition~\ref{def:damage}. The diagnosis points instead to creating independent examples, and $\pi(l)$ already says which tail labels need them and how many.

The extreme-classification family is not covered, because its published results use corpora for which we could not obtain usable text and no prediction files are released for it. The retrieval-augmented audit is one small model on one corpus, run to demonstrate the evaluator, not to characterize language models.

\section{Conclusion}
\label{sec:conclusion}

Macro-F1 over the tail cannot tell whether a rare label was recovered on its own or on the strength of its frequent companions, because it depends only on per-label confusion counts. We decompose per-label recall into companion-supported and independent components and propose a within-label estimand with a co-report rule, and a screening statistic computed from training labels that bounds the inflation of reported recall.

Screening \nScreened{} corpora and auditing the \nEstimableCorpora{} on which the estimand is defined, we find companion dependence wherever it can be measured. No standard long-tail objective reduces it, and none behaves the same way on two corpora, so a tail result is evidence about the benchmark it was measured on. Where the effect moves, it moves with architecture. Tail scores derived from other labels' predictions read as strongly dependent. The screening statistic bounds the inflation of reported recall on every corpus, conservatively, from training labels alone. For collections indexed under large annotation vocabularies, independent recall is the part of a rare descriptor's performance that adds access for a searcher, and it is the part aggregate scores do not report.

Two lessons carry beyond this paper. The intuitive difference-of-F1 construction is dominated by a base-rate artifact and can reverse the sign of the effect, and a lower dependence reading is a correction only when independent recall rises. Future work is to create independent examples for the tail labels the screening statistic identifies, and to extend the audit to extreme classifiers once their predictions are available.

\appendix
\counterwithin{table}{section}
\counterwithin{figure}{section}
\counterwithin{equation}{section}
\renewcommand{\thetable}{\thesection.\arabic{table}}
\renewcommand{\thefigure}{\thesection.\arabic{figure}}
\renewcommand{\theequation}{\thesection.\arabic{equation}}

\section{Supplementary tables}
\label{app:tables}

Three tables referenced in the main text are given here: the per-corpus training setup (Table~\ref{tab:setup}), the bootstrap intervals of the objective contrasts (Table~\ref{tab:objectives-boot}) and the scale of the inflation (Table~\ref{tab:inflation-scale}).

\begin{table}
\caption{The training setup behind every corpus-level figure. Each loss-side run is a label-wise attention classifier on the encoder listed, held fixed across all objectives and arms on that corpus. Length is the input truncation in tokens. Epochs is the training budget, with early stopping on validation loss at a patience of two; a bracketed cap means a run still improving at the budget could continue to it. Tail F1 is the micro-F1 over tail labels of the binary cross-entropy baseline, mean $\pm$ standard deviation across its seeds.}
\label{tab:setup}
\begin{tabular*}{\tblwidth}{@{\extracolsep{\fill}}llrrrl@{}}
\toprule
Corpus & Encoder & Length & Epochs & Objectives & Tail F1 (BCE) \\
\midrule
EURLEX57K & Legal-BERT & 512 & 8 & 7 & $0.168 \pm 0.009$ \\
PubMed MeSH & BiomedBERT & 256 & 5 & 2 & $0.089 \pm 0.008$ \\
Ohsumed & BiomedBERT & 256 & 5 & 7 & $0.075 \pm 0.004$ \\
Wiki10-31K & BERT-base & 512 & 10 (20) & 7 & $0.058 \pm 0.001$ \\
AmazonCat-13K & BERT-base & 256 & 6 (12) & 1 & $0.110 \pm 0.018$ \\
LexGLUE EURLEX & Legal-BERT & 512 & 8 & 1 & $0.525 \pm 0.017$ \\
AAPD & BERT-base & 256 & 8 & 7 & $0.428 \pm 0.011$ \\
arXiv-157 & BERT-base & 256 & 8 & 1 & $0.446 \pm 0.008$ \\
\bottomrule
\end{tabular*}
\end{table}
\begin{table}[t]
\caption{Table~\ref{tab:objectives}'s contrasts under a hierarchical bootstrap that resamples estimable labels and paired seeds together. Each cell is the 95\% interval of $\Delta$, the change in the relative within-label effect against binary cross-entropy. Bold marks an interval that excludes zero and a dagger marks a cell flagged in Table~\ref{tab:objectives}.}
\label{tab:objectives-boot}
\begin{tabular*}{\tblwidth}{@{}lrrr@{}}
\toprule
Objective & Ohsumed & EURLEX57K & Wiki10-31K \\
\midrule
focal & $\mathbf{[+0.012, +0.062]}$\textsuperscript{\dag} & $[-0.200, +0.308]$ & $[-0.133, +0.196]$ \\
asymmetric & $[-0.027, +0.026]$ & $[-0.216, +0.301]$ & $[-0.223, +0.155]$ \\
class-balanced & $[-0.011, +0.050]$ & $[-0.074, +0.501]$\textsuperscript{\dag} & $[-0.113, +0.261]$ \\
distribution-balanced & $[-0.061, +0.102]$ & $[-0.021, +0.537]$\textsuperscript{\dag} & $[-0.390, +0.084]$\textsuperscript{\dag} \\
logit adjustment & $\mathbf{[+0.000, +0.053]}$\textsuperscript{\dag} & $[-0.121, +0.355]$ & $[-0.200, +0.146]$ \\
LDAM & $[-0.021, +0.016]$ & $[-0.118, +0.236]$ & $[-0.048, +0.069]$ \\
\bottomrule
\end{tabular*}
\end{table}
\begin{table}[t]
\caption{Why the measured inflation in Table~\ref{tab:inflation} is small. Per label $R - R_I = \pi(l)\,(R_S - R_I) \leq \pi(l)\,R_S$, so absolute inflation is capped by tail recall however dependent a model is. Relative inflation is the mean of $R - R_I$ over $F$ divided by the mean of $R$ over $F$, the ratio form of $d_{\text{rel}}$. Excess is the share of tail true positives over $F$ that would disappear if supported occurrences were recalled only at the independent rate, $\sum_F |S(l)|\,(R_S - R_I) / \sum_F \mathrm{TP}_l$. Outside $F$ counts the tail labels with test positives but no independent one, on which reported recall cannot be checked for independence at all, with their share of tail labels with any test positive. Inflation over $E$, the labels with both contexts present, equals the mean of $\pi(l)\,d_l$; it excludes labels with no supported positive, whose inflation is zero, and is therefore larger by construction. Binary cross-entropy runs, mean over runs.}
\label{tab:inflation-scale}
\begin{tabular*}{\tblwidth}{@{\extracolsep{\fill}}l@{\hspace{1.6em}}rrr@{\hspace{1.6em}}rr@{\hspace{1.6em}}rr@{}}
\toprule
 & \multicolumn{3}{c}{over $F$} & \multicolumn{2}{c}{outside $F$} & \multicolumn{2}{c}{over $E$} \\
\cmidrule(lr){2-4} \cmidrule(lr){5-6} \cmidrule(l){7-8}
Corpus & absolute & relative & excess & labels & share & $|E|$ & absolute \\
\midrule
EURLEX57K & $0.011$ & $0.037$ & $6.2$\% & 146 & 24\% & 35 & $0.141$ \\
PubMed MeSH & $0.008$ & $0.044$ & $7.6$\% & 1{,}545 & 33\% & 357 & $0.070$ \\
Ohsumed & $0.025$ & $0.156$ & $18.4$\% & 420 & 6\% & 1{,}986 & $0.086$ \\
Wiki10-31K & $0.002$ & $0.014$ & $3.0$\% & 2{,}285 & 22\% & 260 & $0.077$ \\
AmazonCat-13K & $0.008$ & $0.023$ & $3.8$\% & 970 & 15\% & 910 & $0.047$ \\
LexGLUE EURLEX & $0.005$ & $0.009$ & $0.7$\% & 0 & 0\% & 7 & $0.037$ \\
AAPD & $-0.000$ & $-0.000$ & $-0.7$\% & 0 & 0\% & 4 & $-0.003$ \\
arXiv-157 & $0.000$ & $0.000$ & $0.1$\% & 0 & 0\% & 2 & $0.007$ \\
\bottomrule
\end{tabular*}
\end{table}

\section{Mechanism ladders in full}
\label{app:ladders}

Table~\ref{tab:ladders} gives every mechanism-ladder arm measured at three or more runs against its own control. Figure~\ref{fig:ladder} draws the same rows as shifts over the control.

\begin{table*}[t]
\footnotesize
\setlength{\tabcolsep}{4pt}
\caption{Mechanism ladders. Each family is compared against its own control. For the graph, correlation-network, reranking and derivation families the control is the same architecture with the mechanism disabled, holding capacity and training fixed. The sequence-decoding family has no such variant, so its control is the plain BCE baseline and its rows carry a capacity difference the others do not. Figures are the relative within-label effect. sd is the spread across seeds, and $\sigma$ the separation of the arm from its control in standard errors, undefined where either side has fewer than two seeds. Table~\ref{tab:arms} specifies every arm.}
\label{tab:ladders}
\centering

\begin{minipage}[t]{0.48\textwidth}
\begin{tabular*}{\textwidth}[t]{@{\extracolsep{\fill}}llrrr@{}}
\toprule
Corpus & Arm & effect & sd & $\sigma$ \\
\midrule
\multicolumn{5}{@{}l}{\emph{EURLEX57K --- graph}} \\
 & identity\textsuperscript{c} & 0.661 & 0.131 & -- \\
 & co-occurrence & 0.729 & 0.109 & 0.9 \\
 & GATv2 & 0.732 & 0.108 & 0.9 \\
 & dynamic & 0.960 & 0.175 & 3.0 \\
\addlinespace
\multicolumn{5}{@{}l}{\emph{EURLEX57K --- correlation network}} \\
 & diagonal\textsuperscript{c} & 0.654 & 0.103 & -- \\
 & frozen\textsuperscript{c} & 0.561 & 0.075 & -- \\
 & CorNet & 0.618 & 0.099 & 0.2 \\
\addlinespace
\multicolumn{5}{@{}l}{\emph{EURLEX57K --- reranking}} \\
 & seed-only\textsuperscript{c} & 0.752 & 0.126 & -- \\
 & random\textsuperscript{c} & 0.694 & 0.096 & -- \\
 & co-occurrence & 0.646 & 0.152 & 1.0 \\
\addlinespace
\multicolumn{5}{@{}l}{\emph{EURLEX57K --- sequence decoding}} \\
 & BCE\textsuperscript{c} & 0.591 & 0.153 & -- \\
 & SGM & 0.916 & 0.109 & 3.9 \\
 & label-only & 1.228 & 0.318 & 4.0 \\
 & SGM, teacher-forced & 0.921 & 0.427 & 1.6 \\
\addlinespace
\multicolumn{5}{@{}l}{\emph{EURLEX57K --- derivation}} \\
 & gate\textsuperscript{c} & 0.663 & 0.127 & -- \\
 & substitute & 1.427 & 0.158 & 8.4 \\
\midrule
\multicolumn{5}{@{}l}{\emph{Ohsumed --- graph}} \\
 & identity\textsuperscript{c} & 0.316 & 0.013 & -- \\
 & co-occurrence & 0.335 & 0.013 & 2.2 \\
 & GATv2 & 0.349 & 0.016 & 3.4 \\
 & dynamic & 0.345 & 0.025 & 2.2 \\
\addlinespace
\multicolumn{5}{@{}l}{\emph{Ohsumed --- correlation network}} \\
 & diagonal\textsuperscript{c} & 0.358 & 0.028 & -- \\
 & frozen\textsuperscript{c} & 0.365 & 0.011 & -- \\
 & CorNet & 0.375 & 0.013 & 1.5 \\
\addlinespace
\multicolumn{5}{@{}l}{\emph{Ohsumed --- reranking}} \\
 & seed-only\textsuperscript{c} & 0.357 & 0.016 & -- \\
 & random\textsuperscript{c} & 0.363 & 0.022 & -- \\
 & co-occurrence & 0.383 & 0.063 & 0.9 \\
\bottomrule
\end{tabular*}
\end{minipage}%
\hfill
\begin{minipage}[t]{0.48\textwidth}
\begin{tabular*}{\textwidth}[t]{@{\extracolsep{\fill}}llrrr@{}}
\toprule
Corpus & Arm & effect & sd & $\sigma$ \\
\midrule
\multicolumn{5}{@{}l}{\emph{Ohsumed --- sequence decoding}} \\
 & BCE\textsuperscript{c} & 0.367 & 0.021 & -- \\
 & SGM & 0.568 & 0.046 & 10.8 \\
 & label-only & 0.717 & 0.126 & 6.8 \\
 & SGM, teacher-forced & 0.537 & 0.038 & 10.1 \\
\addlinespace
\multicolumn{5}{@{}l}{\emph{Ohsumed --- derivation}} \\
 & gate\textsuperscript{c} & 0.363 & 0.011 & -- \\
 & substitute & 0.916 & 0.101 & 17.1 \\
\midrule
\multicolumn{5}{@{}l}{\emph{PubMed MeSH --- graph}} \\
 & identity\textsuperscript{c} & 0.212 & 0.048 & -- \\
 & co-occurrence & 0.224 & 0.065 & 0.3 \\
 & GATv2 & 0.264 & 0.024 & 2.2 \\
 & dynamic & 0.271 & 0.067 & 1.6 \\
\addlinespace
\multicolumn{5}{@{}l}{\emph{PubMed MeSH --- correlation network}} \\
 & diagonal\textsuperscript{c} & 0.333 & 0.060 & -- \\
 & frozen\textsuperscript{c} & 0.319 & 0.054 & -- \\
 & CorNet & 0.327 & 0.092 & 0.0 \\
\addlinespace
\multicolumn{5}{@{}l}{\emph{PubMed MeSH --- reranking}} \\
 & seed-only\textsuperscript{c} & 0.340 & 0.069 & -- \\
 & random\textsuperscript{c} & 0.312 & 0.036 & -- \\
 & co-occurrence & 0.321 & 0.076 & 0.1 \\
\addlinespace
\multicolumn{5}{@{}l}{\emph{PubMed MeSH --- sequence decoding}} \\
 & BCE\textsuperscript{c} & 0.381 & 0.043 & -- \\
 & SGM & 0.416 & 0.078 & 0.8 \\
 & SGM, teacher-forced & 0.453 & 0.094 & 1.5 \\
\addlinespace
\multicolumn{5}{@{}l}{\emph{PubMed MeSH --- derivation}} \\
 & gate\textsuperscript{c} & 0.319 & 0.057 & -- \\
 & substitute & 0.736 & 0.355 & 2.6 \\
\midrule
\multicolumn{5}{@{}l}{\emph{Wiki10-31K --- sequence decoding}} \\
 & BCE\textsuperscript{c} & 0.960 & 0.098 & -- \\
 & SGM & 0.878 & 0.092 & 1.4 \\
 & label-only\textsuperscript{d} & 0.312 & 0.620 & 2.3 \\
\addlinespace
\multicolumn{5}{@{}l}{\emph{Wiki10-31K --- derivation}} \\
 & gate\textsuperscript{c} & 0.941 & 0.075 & -- \\
 & substitute\textsuperscript{d} & 0.547 & 0.397 & 2.2 \\
\midrule
\multicolumn{5}{@{}l}{\emph{AmazonCat-13K --- derivation}} \\
 & gate\textsuperscript{c} & 0.297 & 0.026 & -- \\
 & substitute & 0.474 & 0.132 & 2.9 \\
\bottomrule
\multicolumn{5}{@{}l}{\footnotesize\textsuperscript{c} control arm. \textsuperscript{d} damage notice (Section~\ref{sec:rule3}).} \\
\end{tabular*}
\end{minipage}

\end{table*}

\section{The screening register}
\label{app:register}

Table~\ref{tab:register} lists every screened corpus with the statistics S1--S4 ask for, all computed from training labels alone. Legal text alone does not produce the phenomenon. ECtHR, with a statutory article list, screens at $0.105$, while EURLEX57K, with a thesaurus, screens at $0.411$.

\begin{table*}[t]
\small
\caption{The screening register. $|D|$ and $|L|$ are training documents and labels. "card." is the mean labels per document. $\bar{\pi}$ is the mean support ratio over the tail at $\tau = 0.5$ with the tail as the bottom half by training frequency. $|C|$ is the mean companion-set size over the tail. Every column is computed from the training labels alone. Corpora sharing a cluster share a document set, a source corpus or a vocabulary, and are not independent observations. Ordered by $\bar{\pi}$.}
\label{tab:register}
\begin{tabular*}{\tblwidth}{@{}lrrrrrl@{}}
\toprule
Corpus & $|D|$ & $|L|$ & card. & $\bar{\pi}$ & $|C|$ & cluster \\
\midrule
RCV1-V2 full & 23{,}149 & 749 & 5.12 & 0.621 & 1.76 & rcv1v2 \\
Ohsumed & 54{,}709 & 14{,}623 & 10.32 & 0.570 & 3.01 & mesh \\
Wiki10-31K & 14{,}146 & 30{,}938 & 18.64 & 0.532 & 14.11 & wiki10\_tags \\
MESINESP2 & 249{,}474 & 22{,}658 & 8.32 & 0.531 & 3.28 & mesh \\
PubMed MeSH & 40{,}000 & 21{,}918 & 12.69 & 0.516 & 5.99 & mesh \\
EURLEX-4K & 15{,}449 & 3{,}956 & 5.32 & 0.420 & 2.06 & eurlex \\
EURLEX57K & 45{,}000 & 4{,}271 & 5.07 & 0.411 & 2.05 & eurlex \\
Chilean dark & 424 & 6 & 1.13 & 0.407 & 0.67 & chileantos \\
RCV1-V2 regions & 23{,}149 & 296 & 1.28 & 0.289 & 0.56 & rcv1v2 \\
AmazonCat-13K & 1{,}186{,}239 & 13{,}330 & 5.04 & 0.254 & 1.09 & amazon \\
RCV1-V2 topics & 23{,}149 & 103 & 3.18 & 0.194 & 0.54 & rcv1v2 \\
Chilean abusive & 1{,}068 & 24 & 1.50 & 0.173 & 0.58 & chileantos \\
Chilean illegal & 551 & 9 & 1.05 & 0.125 & 0.25 & chileantos \\
RCV1-V2 industries & 9{,}644 & 350 & 1.57 & 0.118 & 0.23 & rcv1v2 \\
ECtHR & 9{,}000 & 10 & 1.46 & 0.105 & 0.20 & lexglue\_other \\
AAPD & 53{,}840 & 54 & 2.41 & 0.091 & 0.22 & arxiv \\
LexGLUE EURLEX & 55{,}000 & 100 & 4.42 & 0.082 & 0.18 & eurlex \\
arXiv-157 hier. & 163{,}168 & 142 & 2.97 & 0.024 & 0.07 & arxiv \\
arXiv-157 & 163{,}168 & 130 & 1.37 & 0.017 & 0.03 & arxiv \\
DBpedia & 240{,}942 & 298 & 3.00 & 0.000 & 0.00 & wikipedia \\
GoEmotions & 43{,}410 & 28 & 1.18 & 0.000 & 0.00 & reddit \\
UnfairToS & 5{,}532 & 8 & 0.13 & 0.000 & 0.00 & lexglue\_other \\
Chilean grey & 411 & 9 & 1.32 & 0.000 & 0.00 & chileantos \\
\bottomrule
\end{tabular*}
\end{table*}

\section{Threshold sensitivity}
\label{app:sensitivity}

The results of Section~\ref{sec:results} are reported at one cell of a three-parameter grid. Table~\ref{tab:sensitivity} crosses the companion threshold $\tau$ with the head--tail fraction on every corpus where the aggregate is estimable at the reported cell, and Table~\ref{tab:sensitivity-m} varies the estimability floor $m$.

The reported cell is not a favorable corner. It lies inside the range of its grid on every corpus rather than at either end. The direction does not depend on the thresholds either. Every cell is positive on every corpus. Each entry is a mean over seeds at one $(\tau, \text{tail})$ setting, which separates threshold sensitivity from run-to-run variation. Individual runs do fall slightly below zero on the corpora with the fewest estimable labels, down to $-0.014$ on the corpus whose $|E|$ is seven, which is why Table~\ref{tab:corpus} reports $|E|$ beside every effect.

The floor $m$ changes magnitude and availability, but not sign. Raising $m$ leaves the sign unchanged wherever the effect remains estimable, but moves its size in both directions. At $m = 2$ the effect nearly doubles on PubMed MeSH and falls by two thirds on Wiki10-31K. It also removes the estimate. At $m = 2$ EURLEX57K keeps a single label estimable in both contexts, and at $m = 5$ three of the six corpora keep none. We therefore set $m = 1$, the definedness floor of Definition~\ref{def:partition}. LexGLUE EURLEX reads identically at all three floors, because each of its seven estimable labels already carries five positives in both contexts.

Missing cells indicate entailment. At a narrow tail the surviving labels are rare enough that their companions are entailed by the annotation. The cutoff of Definition~\ref{def:companion} removes them, and the measure reports nothing where a number would come from an empty partition. Where this happens follows from the corpus's label frequencies and, like the screening statistics, is computable before training.

\begin{table}[t]
\caption{Sensitivity of the within-label effect to the estimability floor $m$ of Equation~\ref{eq:within}, at the reported $(\tau, \text{tail})$ cell. Each entry is the mean over the contributing binary cross-entropy runs, with $|E|$ beneath it. A dash marks a floor at which the corpus carries fewer than five labels estimable in both contexts, where Section~\ref{sec:estimand} reports the quantity as undefined. Raising $m$ does not change the sign anywhere it remains estimable. It changes the magnitude, and on three of the six corpora it removes the estimate entirely.}
\label{tab:sensitivity-m}
\begin{tabular*}{\tblwidth}{@{}lrrr@{}}
\toprule
Corpus & $m = 1$ & $m = 2$ & $m = 5$ \\
\midrule
Ohsumed & $+0.131$ & $+0.129$ & $+0.107$ \\
\quad\footnotesize $|E|$ & \footnotesize 1986 & \footnotesize 1569 & \footnotesize 793 \\[2pt]
EURLEX57K & $+0.280$ & -- & -- \\
\quad\footnotesize $|E|$ & \footnotesize 35 & \footnotesize -- & \footnotesize -- \\[2pt]
Wiki10-31K & $+0.133$ & $+0.048$ & -- \\
\quad\footnotesize $|E|$ & \footnotesize 260 & \footnotesize 7 & \footnotesize -- \\[2pt]
PubMed MeSH & $+0.110$ & $+0.197$ & -- \\
\quad\footnotesize $|E|$ & \footnotesize 357 & \footnotesize 18 & \footnotesize -- \\[2pt]
AmazonCat-13K & $+0.084$ & $+0.105$ & $+0.119$ \\
\quad\footnotesize $|E|$ & \footnotesize 910 & \footnotesize 339 & \footnotesize 14 \\[2pt]
LexGLUE EURLEX & $+0.073$ & $+0.073$ & $+0.073$ \\
\quad\footnotesize $|E|$ & \footnotesize 7 & \footnotesize 7 & \footnotesize 7 \\[2pt]
\bottomrule
\end{tabular*}
\end{table}
\begin{table}[t]
\caption{Sensitivity of the within-label effect to the three swept thresholds. Each row is one corpus: the reported cell is $\tau = 0.5$ with the tail as the bottom half by training frequency, and the range covers every combination of $\tau \in \{0.3, 0.4, 0.5, 0.6, 0.7\}$ and tail fraction $\in \{0.3, 0.5, 0.7\}$ in which the aggregate is estimable. Cells is the number estimable out of those attempted: where the effect becomes undefined it does so because $|E|$ falls below the floor. The floor is a fact about the corpus rather than instability in the estimate.}
\label{tab:sensitivity}
\begin{tabular*}{\tblwidth}{@{}lrrrr@{}}
\toprule
Corpus & reported & min & max & cells \\
\midrule
Ohsumed & $+0.130$ & $+0.112$ & $+0.137$ & 10/15 \\
EURLEX57K & $+0.280$ & $+0.247$ & $+0.491$ & 13/15 \\
Wiki10-31K & $+0.133$ & $+0.111$ & $+0.175$ & 11/15 \\
PubMed MeSH & $+0.110$ & $+0.063$ & $+0.145$ & 13/15 \\
AmazonCat-13K & $+0.084$ & $+0.075$ & $+0.155$ & 15/15 \\
LexGLUE EURLEX & $+0.073$ & $+0.037$ & $+0.130$ & 8/15 \\
\bottomrule
\end{tabular*}
\end{table}

\section{The base-rate null}
\label{app:null}

The null column of Table~\ref{tab:corpus} asks how large the difference-of-F1 gap of Equation~\ref{eq:diff-f1} would be for a model with no companion dependence, one whose output does not depend on companion presence given the truth. For each tail label we take the model's own recall and false-positive rate, apply them in both contexts of the real partition, and recompute the gap. Expected true positives are split between $S(l)$ and $I(l)$ in proportion to $|S(l)|$ and $|I(l)|$, and expected false positives between companion-carrying and companion-free negatives in proportion to their counts. Because each label keeps its own rates, the null permits labels to differ from each other and forbids only within-label context dependence, the more conservative of the two nulls we computed. The null share is the null gap as a percentage of the observed gap, averaged over five BCE seeds. A share above 100\% means the base-rate artifact alone exceeds the observed gap. A negative share arises where the observed gap and the null gap have opposite signs: on Wiki10-31K, AAPD and arXiv-157 the observed gap is positive, conditioning appearing to raise tail F1, while the null predicts a negative gap, so the artifact there pushes against the reading rather than producing it.

\section{The retrieval-augmented prompt}
\label{app:prompt}

The generator receives the following prompt, with the document truncated to a fixed character budget and the 40 retrieved label names, at temperature zero with reasoning disabled.

\begin{small}
\begin{verbatim}
You are indexing a MEDLINE abstract with MeSH subject headings.

Here are candidate entries, one per line. Choose only those that genuinely
apply to the document. Choose only from this list. Most have 10 to 20.

CANDIDATES:
{40 label names}

DOCUMENT:
{document}

Reply with the applicable entries, one per line, and nothing else.
\end{verbatim}
\end{small}

\section*{Data availability}

The evaluator and the per-label confusion counts from which every reported estimand is computed are made available in a repository on GitHub that will be public on publication. Per-run prediction files are retained and available on request. The corpora are third-party and are not redistributed. Every one is public except the Chilean Terms of Service collection, which is available from its authors on request.

\section*{Acknowledgments}

This research was supported by Pontificia Universidad Católica de Valparaíso, Vice Rector's Office for Research, Creation and Innovation, within the program ``DI Asociativo Equipamiento 2026'' (044.231/2026).

\end{document}